\documentclass[letterpaper, 10 pt, conference]{ieeeconf}

\IEEEoverridecommandlockouts                              
\usepackage[utf8]{inputenc}

\usepackage{graphicx}
\usepackage{epsfig}
\graphicspath{{./Figures/}}
\usepackage{amsmath}
\usepackage{amssymb}
\usepackage{ulem}
\usepackage{xcolor}
\usepackage{url}
\usepackage{caption}
\usepackage{subcaption}

\usepackage[ruled, linesnumbered]{algorithm2e}

\newtheorem{remark}{Remark}

\newtheorem{property}{Property}

\DeclareMathOperator*{\argmax}{argmax}

\title{\LARGE \bf
Bernstein polynomial-based transcription method for solving optimal trajectory generation problems
}

\author{Calvin Kielas-Jensen$^{1}$ and Venanzio Cichella$^{1}$
\thanks{$^{1}$Calvin Kielas-Jensen and Venanzio Cichella are with the Mechanical Engineering department,
        University of Iowa,
        Iowa City, IA 52242, USA
        {\tt\small calvin-kielas-jensen@uiowa.edu}, {\tt\small venanzio-cichella@uiowa.edu}}%
}

\begin{document}

\maketitle
\thispagestyle{empty}
\pagestyle{empty}

\begin{abstract}
    This paper presents a method and an open-source implementation, Bernstein/B\'ezier Optimal Trajectories (BeBOT), for the generation of trajectories for autonomous system operations. The proposed method is based on infinite dimensional optimal control formulations of trajectory generation problems. By approximating the trajectories using Bernstein polynomials, these problems can be transcribed as nonlinear programming problems, which can then be solved using off-the-shelf solvers. Bernstein polynomials possess favorable geometric properties that enable the trajectory planner to efficiently evaluate and enforce constraints along the vehicles' trajectories, including maximum speed and angular rates, minimum distance between trajectories and between the vehicles and obstacles. By virtue of these properties, feasibility and safety constraints typically imposed in autonomous vehicle operations can be enforced and guaranteed independently on the order of the polynomials. Thus, the trajectory generation algorithm can efficiently generate feasible and collision-free trajectories, and can be deployed for real-time safety critical applications in complex environments and for multiple vehicle missions.
\end{abstract}

\section{Introduction}

The field of autonomous navigation has exploded in the past decade. Significant progress has been made in self driving vehicles, bringing them one step closer to reality \cite{milford2019self}. Precision agriculture utilizes autonomous aerial vehicles to monitor crops and spray pesticides \cite{mogili2018review}, and development in autonomous weed pulling robots may reduce or eliminate the need for potentially harmful pesticides \cite{koerhuis_2018}. Companies such as Amazon, Starship, and Zipline have already begun making autonomous deliveries \cite{ackerman_2015,ackerman_koziol_2019,lardinois_2019}. There is even a mission underway to fly the first autonomous aerial vehicle on a different planet \cite{johnson_hautaluoma_2020}. This progress has led to high demand for computationally efficient algorithms that allow safe and optimal trajectories to be planned for groups of autonomous vehicles. Our proposed method aims to accomplish these tasks by formulating the optimal trajectory generation problem as a nonlinear programming problem and exploiting the useful features of Bernstein polynomials.

Most techniques for planning and control of autonomous systems fall into one of two categories: closed-loop methods or open-loop methods. Closed-loop methods, sometimes referred to as feedback or reactive methods, use the current state knowledge to determine, in real time, what the next control value should be. On the other hand, open-loop methods determine control values or motion trajectories out to a specified time horizon with the use of the system's model.

One common closed-loop technique that originally stemmed from maze solving algorithms is the bug algorithm. The bug algorithm, e.g. \cite{lumelsky1987path,kamon1997sensory}, uses local knowledge of the environment and a global goal to either follow a wall or move in a straight line towards the goal. This algorithm can be implemented on very simple devices due to typically requiring only two tactile sensors. However, it does not provide optimal paths and does not generally take the system's dynamics into account. For a review and comparison of bug algorithms, the reader is referred to \cite{mcguire2018comparative}.

Rather than working on an agent's positions, the velocity obstacle (VO) algorithm uses relative velocities between the agent and obstacles to determine trajectories which will avoid collisions. The original term velocity obstacle was presented in \cite{fiorini1993motion}. Variations on the VO method include Common Velocity Obstacle \cite{abe2001collision}, Nonlinear Velocity Obstacles \cite{large2002using}, and Generalized Velocity Obstacles \cite{wilkie2009generalized}. Other relevant closed-loop methods use artificial potential fields, which leverage a potential function providing attractive forces towards the goal and repulsive forces away from obstacles \cite{khatib1986real,chen2016uav,orozco2019mobile}.

Among the advantages of closed-loop methods are fast computation and the ability to react to changing environments and unforeseen events. Furthermore, theoretical tools aimed at deriving safety guarantees of closed-loop methods are fairly well developed, and are mostly rooted in nonlinear systems analysis, and robust and adaptive control. Despite these benefits, closed-loop methods are difficult to employ for multiple vehicle teams. They also generally lack the capability of presenting a human operator with a predicted trajectory and act rather like a black box.

In contrast to closed-loop methods, open-loop methods can generate solutions in one-shot for the whole mission time, and are therefore able to present an operator with an intuitive representation of the future trajectory. This representation is typically shown as a 2D or 3D path and may also include speed, acceleration, and higher derivatives of the vehicle's motion. Randomized algorithms such as probabilistic roadmaps (PRMs) \cite{kavraki1996probabilistic} and rapidly exploring random trees (RRT,RRT*) \cite{lavalle2001randomized,karaman2010incremental} randomly sample the work space to reduce computational complexity. PRMs randomly sample feasible regions within the work space to construct a dense graph. A graph-based solver can then be used to determine the optimal route. RRTs compute trajectories by using directed sampling to build trees. This approach can find feasible solutions in situations involving both a high number of constraints and high dimensional search spaces. Unfortunately, random sampling algorithms may be difficult to use in real-time applications due to computational complexity and exploring regions that will not lead to a solution.

Similar to PRMs, other graph-based approaches aim to efficiently build and then search a graph. Cell decomposition methods, e.g. \cite{sleumer1999exact,cai2009information}, build a graph of their environment by recursively increasing the resolution of areas of interest resulting in a few large nodes of open space and many small nodes near obstacles. Once a graph has been built, a graph solver can be used. A popular graph solver is the A* algorithm \cite{hart1968formal} which is an extension of Dijkstra's algorithm that uses a heuristic function to improve the search speed.

In addition to graph-based representations of trajectories, polynomial approximation methods can be used as well. In \cite{mellinger2011minimum} trajectories are represented as piecewise polynomial functions and are generated in a manner that minimizes their snap. In TrajOpt \cite{TrajOpt} a sequential quadratic program is solved to generate optimal polynomial trajectories while performing continuous time collision checking.

Other open-loop methods include CHOMP \cite{ratliff2009chomp,zucker2013chomp}, STOMP \cite{kalakrishnan2011stomp}, and HOOP \cite{hoop}. In CHOMP, infeasible trajectories are pulled out of collisions while simultaneously smoothing the trajectories using covariant gradient descent. STOMP adopts a similar cost function to that found in CHOMP but generalizes to cost functions whose gradients are not available. This is done by stochastically sampling noisy trajectories. HOOP utilizes a problem formulation which computes vehicle trajectories in two steps. In the first step, a path is planned quickly by considering only the robot's kinematics. The second step then refines this trajectory into a higher order piecewise polynomial using a quadratic program.

Open-loop methods provide useful tools for dealing with high dimensional problems such as multiple vehicles and several constraints. They are also capable of producing trajectories that accomplish multiple goals. However, due to the curse of dimensionality, the computational complexity of open-loop methods grows significantly with the number of vehicles, constraints, and goals. For the most part, motion planning methods trade optimality and/or safety for computational speed. Our goal is to introduce a method that mitigates this trade-off, and that provides provably safe solutions for high dimensional problems while retaining the computational efficiency of low-order trajectory planning algorithms. This is achieved by exploiting the useful features of Bernstein polynomials.

The Bernstein basis was originally introduced by Sergei Natanovich Bernstein (1880-1968) in order to provide a constructive proof of the Weierstrass theorem. Bernstein polynomials were not widely used until the advent of digital computers due to their slow convergence as a function approximation. Widespread adoption eventually occurred when it was realized that the coefficients of Bernstein polynomials could be intuitively manipulated to change the shape of curves described by these polynomials. In the 1960s, two French automotive engineers became interested in using this idea: Paul de Faget de Casteljau and Pierre \'{E}tienne B\'{e}zier.

Designing complex shapes for automobile bodies by sculpting clay models proved to be a time consuming and expensive process. To combat this, de Casteljau and B\'{e}zier sought to develop mathematical tools that would allow designers to intuitively construct and manipulate complex shapes. Due to de Casteljau publishing most of his research internally at his place of employment, B\'{e}zier's name became more widely associated with Bernstein polynomials, frequently referred to as B\'{e}zier curves. Building on existing research and modern technology, Bernstein polynomials provide several useful properties for many fields.

The Bernstein basis provides numerical stability \cite{farouki1996optimal}, as well as useful geometric properties and computationally efficient algorithms that can be used to derive and implement efficient algorithms for the computation of trajectory bounds, trajectory extrema, minimum temporal and spatial separation between two trajectories and between trajectories and obstacles, and collision detection. Bernstein polynomials also allow the representation of continuous time trajectories using low-order approximations.

Our method for trajectory generation builds upon \cite{VenOptBern,cichella2018bernstein,cichella2020optimal}, where Bernstein polynomials were introduced as tools to approximate nonlinear optimal control problems with provable convergence guarantees. This paper also extends the preliminary work shown in \cite{kielas2019bebot} by providing an open-source implementation of algorithms for Bernstein polynomials, as well as examples that illustrate their use for trajectory generation applications.

We formulate the motion planning problem as an infinite dimensional optimal control problem. By parameterizing the trajectories as Bernstein polynomials, this problem can then be transcribed into a nonlinear programming problem using, which can then be solved using an off-the-shelf solution such as MATLAB's \textit{fmincon}, SciPy Python's \textit{minimize}, or SNOPT.

The paper is structured as follows. In Section \ref{sec:properties} we introduce the Bernstein polynomials and their properties. In Section \ref{sec:algorithms} we present computationally efficient algorithms for the computation of state and input constraints typical of trajectory generation applications. In Section \ref{sec:examples} we demonstrate the efficacy of these algorithms through several numerical examples. The paper ends with Section \ref{sec:conclusions}, which draws some conclusions. A Python implementation of the properties and algorithms presented, as well as the scripts used to generate the plots and examples found throughout this paper, can be found on our GitHub webpage, \cite{BeBOT}.

In what follows, vectors are denoted by bold letters, e.g. $\mathbf{p} =[p_x \, , \, p_y]^\intercal$. $||\cdot||$ denotes the Euclidean norm (or magnitude), e.g., $||\mathbf{p}|| = \sqrt{p_x^2 + p_y^2}$.

\section{Bernstein Polynomials} \label{sec:properties}

This section examines the properties of Bernstein polynomials and rational Bernstein polynomials that are relevant to this paper.

A $1$-dimensional, $n$th order Bernstein polynomial, ${C}_n(t)$, is defined as
\begin{equation} \label{eq:BezCurve1dim}
    {C}_n(t) = \sum_{i=0}^n{{P}_{i,n}B_{i,n}\left(t\right)}, \quad t \in [t_0, t_f] \, ,
\end{equation}
where ${P}_{i,n} \in \mathbb{R}$ is the $i$th Bernstein coefficient and $B_{i,n}(t)$ is the Bernstein polynomial basis defined as
$$
    B_{i,n}(t) = \binom{n}{i} \frac{ (t-t_0)^i (t_f-t)^{n-i} }{ (t_f - t_0)^n },
$$
for all $i = 0, \dots, n$, where
$$
    \binom{n}{i} = \frac{n!}{i!(n-i)!}
$$
is the binomial coefficient.

Similarly, an $N$-dimensional, $n$th order Bernstein polynomial, $\mathbf{C}_n(t)$, is defined as
\begin{equation} \label{eq:BezCurve}
    \mathbf{C}_n(t) = \sum_{i=0}^n{\mathbf{P}_{i,n}B_{i,n}\left(t\right)}, \quad t \in [t_0, t_f] \, ,
\end{equation}
with $\mathbf{P}_{i,n} \in \mathbb{R}^{N}$. Bernstein polynomials can be used to describe 2D (or 3D) spatial curves. In this case, Bernstein polynomials are often referred to as B{\'e}zier curves. While B{\'e}zier's original work did not explicitly use the Bernstein basis \cite{bezier1966definition,bezier1967definition}, it was later shown that the original formulation is equivalent to the Bernstein form polynomial \cite{forrest1972interactive}.

A 1-dimensional, $n$th order \textit{rational} Bernstein polynomial, $R_n(t)$, is defined as
\begin{equation}\label{eq:rational}
    R_n(t) =
    \frac{\sum_{i=0}^n{P_{i,n} {w}_{i,n} B_{i,n} \left(t\right)}}
         {\sum_{i=0}^n{{w}_{i,n} B_{i,n} \left(t\right)}}, \quad
         t \in [t_0, t_f],
\end{equation}
where ${w}_{i,n} \in \mathbb{R}$, $i=0,\ldots,n$, are referred to as weights.

We provide a review of relevant properties of (rational) Bernstein polynomials, which will be used in the remainder of this paper.
For an extensive review on Bernstein polynomials the reader is referred to \cite{farouki}.

\begin{property} \label{prop:convexhull}
\textit{Convex Hull}

The Bernstein polynomial introduced in \eqref{eq:BezCurve}, as well as the rational Bernstein polynomial in \eqref{eq:rational} (provided that $w_{i, n} > 0, i = 0, \dots, n$), lie within the convex hull defined by their Bernstein coefficients. An illustrative example of this property is depicted in Figure \ref{fig:cvxhull} which shows a 2-dimensional Bernstein polynomial within its convex hull.

Given the $1$-dimensional Bernstein polynomial introduced in \eqref{eq:BezCurve1dim}, the convex hull property implies
\begin{equation}\label{eq:convexhull}
    \min_{k \in \{0,\ldots,n \}} {P}_k \leq {C}_n(t) \leq \max_{k \in \{0,\ldots,n \}} {P}_k \, , \,
    \forall t \in [t_0, t_f].
\end{equation}

\begin{figure}
    \centering
    \includegraphics[scale=0.25]{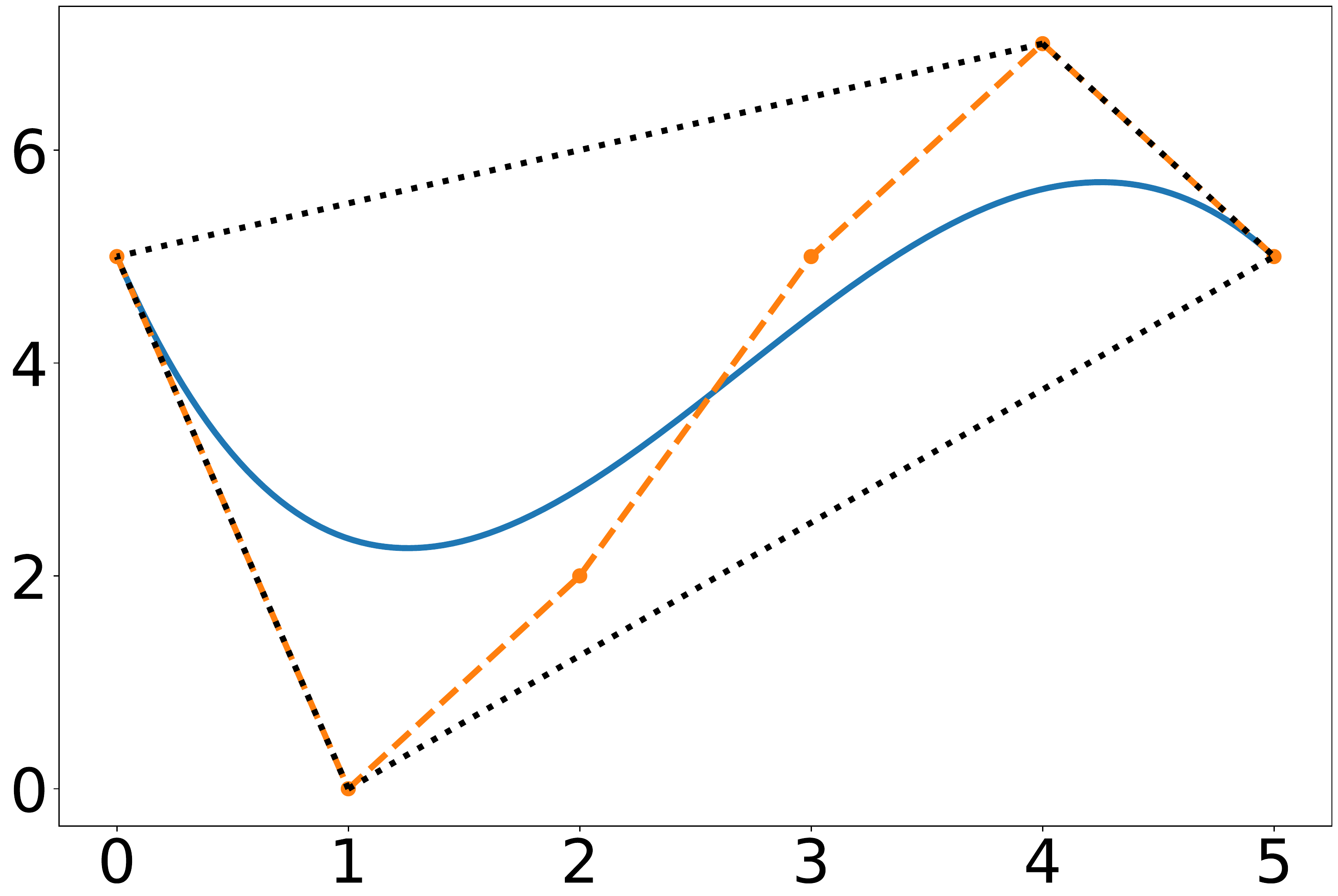}
    \caption{Convex hull of a Bernstein polynomial where the solid line is the curve, the black dotted line is the convex hull, and the points connected by dashed lines are the Bernstein coefficients.}
    \label{fig:cvxhull}
\end{figure}

\end{property}

\begin{property} \label{prop:endpts}
\textit{End Point Values}

The first and last Bernstein coefficients of the Bernstein polynomial introduced in \eqref{eq:BezCurve}, as well as the rational Bernstein polynomial in \eqref{eq:rational}, are their endpoints, i.e., $$\mathbf{C}_n(t_0) = \mathbf{P}_0 \quad \text{and} \quad \mathbf{C}_n(t_f) = \mathbf{P}_n.$$ Furthermore, the first derivative of the Bernstein polynomial at the end points can be easily computed as follows

\begin{equation} \label{eq:derinitpoint}
\begin{split}
&    \dot{\mathbf{C}}_n(t_0) = \frac{n}{t_f - t_0}(\mathbf{P}_{1,n} - \mathbf{P}_{0,n}),
\\
&    \dot{\mathbf{C}}_n(t_f) = \frac{n}{t_f - t_0}(\mathbf{P}_{n,n} - \mathbf{P}_{n-1,n}).
\end{split}
\end{equation}
Similarly, for a rational Bernstein polynomial we have
\begin{equation} \label{eq:ratderinitpoint}
     \dot{\mathbf{C}}_n(t_0) = \frac{n w_1}{(t_f - t_0) w_0}(\mathbf{P}_{1,n} - \mathbf{P}_{0,n}),
 \end{equation}
and
\begin{equation} \label{eq:ratderfinpoint}
     \dot{\mathbf{C}}_n(t_f) = \frac{n w_{n-1}}{(t_f - t_0) w_n}(\mathbf{P}_{n,n} - \mathbf{P}_{n-1,n}).
 \end{equation}
\end{property}

\begin{remark}
Property \ref{prop:endpts} and Equation \eqref{eq:derinitpoint} can be easily extended to compute higher order derivatives of Bernstein polynomials at the end points.
\end{remark}

\begin{property} \label{prop:derivatives}
\textit{Derivatives}

The derivative of the Bernstein polynomial introduced in \eqref{eq:BezCurve} is an $(n-1)$th order Bernstein polynomial
\begin{equation}
    \dot{\mathbf{C}}_{n-1}(t) =
    \sum_{i=0}^{n-1}{\mathbf{P}^\prime_{i,n-1}B_{i,n-1}(t)},
\end{equation}
with vector of Bernstein coefficients $\mathbf{P}^\prime_{n-1} = [\mathbf{P}^\prime_{0,n-1} , \ldots, \mathbf{P}^\prime_{n-1,n-1}]$ given by
$$\mathbf{P}_{n-1}^\prime = \mathbf{P}_{n} \mathbf{D}_n,$$
where
\begin{equation}
\mathbf{D}_n =   \frac{n}{t_f - t_0} \begin{bmatrix}
-1    & 0      & \cdots & 0       \\
1      & \ddots & \ddots & \vdots  \\
0      & \ddots & \ddots & 0       \\
\vdots & \ddots & \ddots & -1      \\
0      & \cdots & 0      & 1
\end{bmatrix} \in \mathbb{R}^{n+1 \times n}.
\end{equation}

\end{property}

\begin{property} \label{prop:integrals}
\textit{Integrals}

The definite integral of a Bernstein polynomial can be found from its coefficients,
\begin{equation}
    \int_{t_0}^{t_f} \mathbf{C}_n(t) dt = \frac{t_f - t_0}{n+1} \sum_{i=0}^n \mathbf{P}_{i, n} \, .
\end{equation}

\end{property}

\begin{property} \label{prop:decast}
\textit{The de Casteljau Algorithm}

The de Casteljau algorithm computes a Bernstein polynomial defined over an interval $[t_0,t_f]$ at any given $t_{\textit{div}} \in [t_0,t_f]$. Moreover, it can be used to split a single Bernstein polynomial into two independent Bernstein polynomials.
Given the Bernstein polynomial introduced in \eqref{eq:BezCurve} with the vector of coefficients $\mathbf{P}_{n}=[\mathbf{P}_{0,n} \, , \, \ldots \, , \, \mathbf{P}_{n,n}]$, and a scalar $t_{\textit{div}} \in [t_0, t_f]$, the Bernstein polynomial at $t_{\textit{div}}$ is computed using the following recursive relationship:
$$
\mathbf{P}_{i,n}^{0} = \mathbf{P}_{i,n}\, , \quad i = 0,\ldots,n \, ,
$$
$$
\mathbf{P}_{i,n}^{j} = \frac{t_f - t_{\textit{div}}}{t_f - t_0} \mathbf{P}_{i,n}^{j-1} + \frac{t_{\textit{div}} - t_0}{t_f - t_0} \mathbf{P}_{i+1,n}^{j-1} \, ,
$$
with $i = 0,\ldots,n-j \, ,$ and $j = 1,\ldots,n$. Then, the Bernstein polynomial evaluated at $t_{\textit{div}}$ is given by
$
\mathbf{C}_n(t_{\textit{div}}) = \mathbf{P}_{0,n}^{n} \, .
$
Note that the superscript here signifies an index and not an exponent.
Moreover, the Bernstein polynomial can be subdivided at $t_{\textit{div}}$ into two $n$th order Bernstein polynomials with coefficients
$$
\mathbf{P}_{0,n}^{0},\mathbf{P}_{0,n}^{1},\ldots,\mathbf{P}_{0,n}^{n} \quad \text{and} \quad
\mathbf{P}_{0,n}^{n},\mathbf{P}_{1,n}^{n-1},\ldots,\mathbf{P}_{n,n}^{0} \, .
$$

A geometric example of a $3$rd order Bernstein polynomial being split at $t_{div} = 0.5$ using the de Casteljau is shown in Figure \ref{fig:decast}. Note that at the final iteration only a single point remains, $\mathbf P_{0, n}^{3}$, and lies on the original curve. The points $\{\mathbf P_{0, n}^{0}, \mathbf P_{0, n}^{1}, \mathbf P_{0, n}^{2}, \mathbf P_{0, n}^{3}\}$ become the Bernstein coefficients of the left curve and the points $\{\mathbf P_{0, n}^{3}, \mathbf P_{1, n}^{2}, \mathbf P_{2, n}^{1}, \mathbf P_{3, n}^{0}\}$ become the coefficients of the right curve.

\begin{figure}
    \centering
    \includegraphics[scale=0.22]{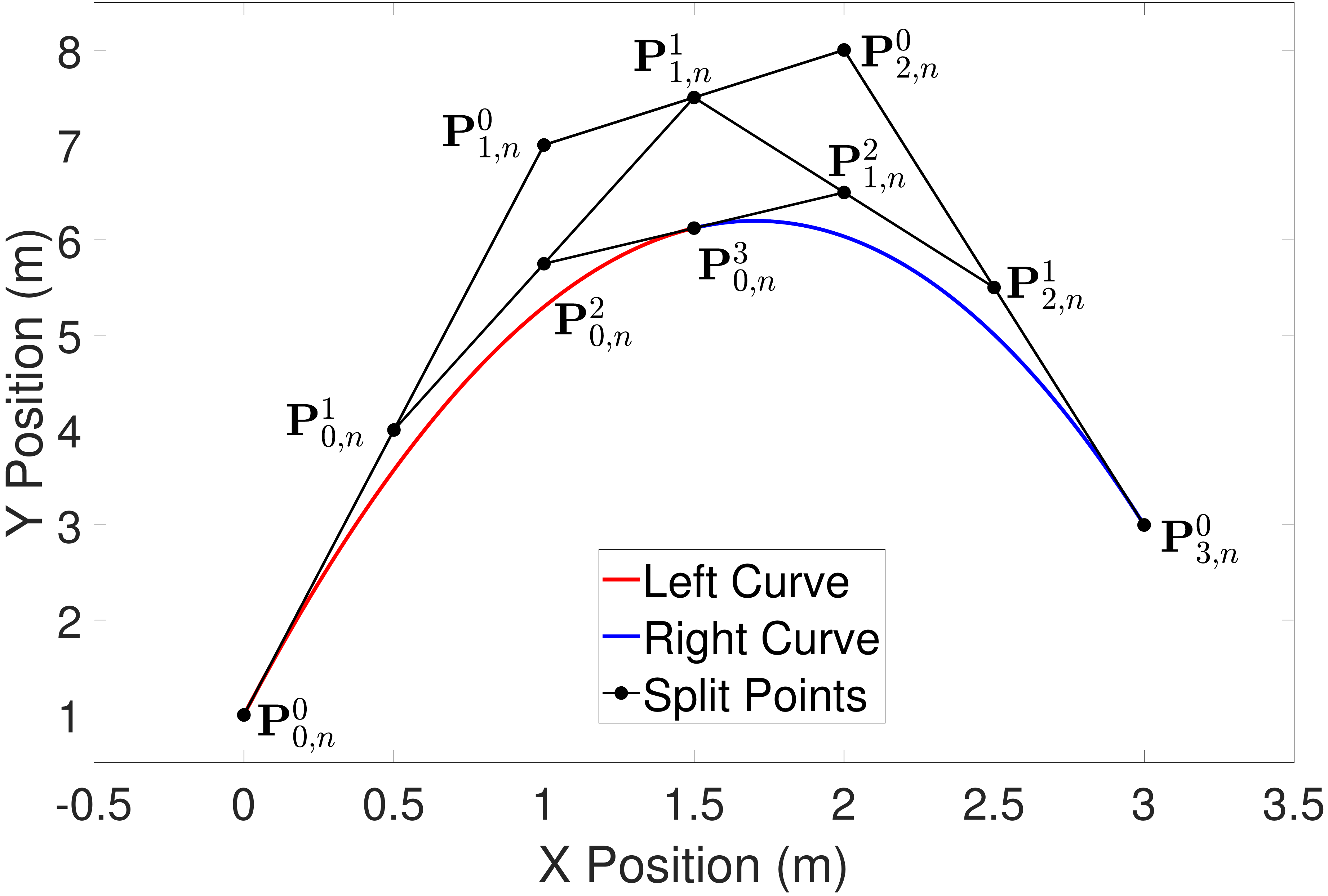}
    \caption{Geometric example of the de Casteljau algorithm splitting a Bernstein polynomial at $t_{div} = 0.5$. The original curve is defined by the Bernstein coefficients $\{\mathbf P_{0, n}^{0}, \dots, \mathbf P_{3, n}^{0}\}$, the left hand curve is shown in red, the right hand curve is shown in blue, and the superscript corresponds to the current iteration of the algorithm.}
    \label{fig:decast}
\end{figure}

\end{property}

\begin{property} \label{prop:elevation}
\textit{Degree Elevation}

The degree of a Bernstein polynomial can be elevated by one using the following recursive relationship
$$
\mathbf{P}_{i, n+1} = \frac{i}{n+1} \mathbf{P}_{i-1, n} + \left( 1 - \frac{i}{n+1} \right) \mathbf{P}_{i, n},
$$
where $\mathbf{P}_{0, n+1} = \mathbf{P}_{0, n}$ and $\mathbf{P}_{n+1, n+1} = \mathbf{P}_{n, n}$.
Furthermore, any Bernstein polynomial of degree $n$ can be expressed as a Bernstein polynomial of degree $m$, $m > n$. The vector of coefficients of the degree elevated Bernstein polynomial, namely $\mathbf{P}_m = [P_{0,m}, \dots, P_{m, m}]$, can be calculated as
$$
    \mathbf{P}_{m} = \mathbf{P}_{n} \mathbf{E},
$$
where $\mathbf{E} = \{e_{j,k}\} \in \mathbb{R}^{(n+1) \times (m+1)}$ is the degree elevation matrix with elements given by
\begin{equation} \label{eq:degelevmatrix}
    e_{i, i+j} = \frac{\binom{m-n}{j} \binom{n}{i}}{\binom{m}{i+j}} \, ,
\end{equation}
where $i = 0, \dots, n$ and $j = 0, \dots, m - n$, all other values in the matrix are zero, and $\mathbf{P}_n = [P_{0, n}, \dots, P_{n, n}]$ is the vector of Bernstein coefficients of the curve being elevated (see \cite{lee1997distance}).
By degree elevating a Bernstein polynomial, the coefficients converge to the polynomial, i.e.
\begin{equation} \label{eq:degelevconv}
    \max_i \left|P_{i,m} - C_{m}\left(\frac{i}{m}(t_f-t_0)+t_0\right)\right| \leq \frac{k}{m} ,
\end{equation}
where $k$ is a positive constant independent of $m$ (see \cite{Prautzsch1994}).
\end{property}

\begin{property} \label{prop:arithmetic}
\textit{Arithmetic Operations}

Arithmetic operations of Bernstein polynomials require that both curves are on the same time interval $[t_0, t_f]$. If they are not on the same time interval, the de Casteljau algorithm can be used to split them at an intersecting time interval.

The sum (difference) of two polynomials of the same order can be performed by simply adding (subtracting) their Bernstein coefficients. If the Bernstein polynomials are not of the same order, Degree Elevation (Property \ref{prop:elevation}) may be used to increase the order of the lower order Bernstein polynomial.

Let $F_m(t)$ and $G_n(t)$ be two 1-dimensional Bernstein polynomials of degree $m$ and $n$, respectively, with Bernstein coefficients $X_{0,m}, \dots, X_{m,m}$ and $Y_{0,n}, \dots, Y_{n,n}$. The product $C_{m+n}(t)=F_m(t)G_n(t)$ is a Bernstein polynomial of degree $(m+n)$ with coefficients $P_{k,m+n}$, $k\in\{0,\ldots,m+n\}$ given by
\begin{equation}
    P_{k,m+n} = \sum_{j=\max(0,k-n)}^{\min(m,k)} \frac{\binom{m}{j} \binom{n}{k-j}}{\binom{m+n}{k}} X_{j, m} Y_{k-j, n}.
\end{equation}

The ratio between two $1$-dimensional Bernstein polynomials, $F_n(t)$ and $G_n(t)$, with coefficients $X_{0,n}, \dots, X_{n,n}$ and $Y_{0,n}, \dots, Y_{n,n}$, i.e., $R_n(t)=F_n(t)/G_n(t)$, can be expressed as a rational Bernstein polynomial as defined in \eqref{eq:rational}, with coefficients and weights
$${P}_{i,n} = \frac{{X}_{i,n}}{{Y}_{i,n}} \, , \quad {w}_{i,n} = {Y}_{i,n}, $$ respectively.
\end{property}

\begin{property} \label{prop:ratrecursive}
\textit{Rational Recursive Algorithm}

The de Casteljau algorithm can be extended to rational Bernstein polynomials (see \cite{farin1983algorithms}). Letting
\begin{equation}
    w_{i, n}^{r}(t) = \sum_{j=0}^{r} w_{i+j, n} B_{i, r}(t),
\end{equation}
we can determine a point on an $n$th order rational Bernstein polynomial using the following recursive equation
\begin{equation}
	\begin{aligned}
	    P_{i, n}^{r}(t) = &\left( \frac{t_f - t}{t_f - t_0} \right)\frac{w_{i, n}^{r-1}(t)}{w_{i, n}^{r}(t)} P_{i, n}^{r-1}(t) \\
	    &+ \left( \frac{t - t_0}{t_f - t_0} \right) \frac{w_{i+1, n}^{r-1}(t)}{w_{i, n}^{r}(t)} P_{i+1, n}^{r-1}(t).
	\end{aligned}
\end{equation}
Moreover, the recursive relationship can be used to split the rational Bernstein polynomial into two $n$th order rational Bernstein polynomials with weights
$$
w_{0,n}^{0}, w_{0,n}^{1}, \ldots, w_{0,n}^{n} \quad \text{and} \quad
w_{0,n}^{n}, w_{1,n}^{n-1}, \ldots, w_{n,n}^{0} \, ,
$$
and coefficients
$$
P_{0,n}^{0}, P_{0,n}^{1}, \ldots, P_{0,n}^{n} \quad \text{and} \quad
P_{0,n}^{n}, P_{1,n}^{n-1}, \ldots, P_{n,n}^{0} \, .
$$
\end{property}

\begin{property} \label{prop:ratelev}
\textit{Rational Degree Elevation}

Degree elevation can be extended to rational Bernstein polynomials (see \cite{farin1983algorithms}). Given an $n$th order rational Bernstein polynomial, the identical rational Bernstein polynomial of order $(n+1)$ can be determined. The weights are found using
$$
w_{i, n+1} = \frac{i}{n+1} w_{i-1, n} + \left(1 - \frac{i}{n+1} \right) w_{i, n}.
$$
where $w_{0, n+1} = w_{0, n}$ and $w_{n+1, n+1} = w_{n, n}$. The Bernstein coefficients are found using
$$
P_{i, n+1} = \frac{ \frac{i}{n+1} w_{i, n} P_{i, n} + \left( 1 - \frac{i}{n+1} \right) w_{i+1, n} P_{i+1, n} }{ \frac{i}{n+1} w_{i, n} + \left( 1 - \frac{i}{n+1} \right) w_{i+1, n} }.
$$

\end{property}

\subsection{Numerical Examples in 2D}

Here we will examine several numerical examples using the properties of Bernstein polynomials in 2D. All the plots presented can be generated using the example code available at \cite{BeBOT}.

\begin{figure*}
\centering
\begin{subfigure}[t]{0.49\textwidth}
    \centering
    \includegraphics[scale=0.2]{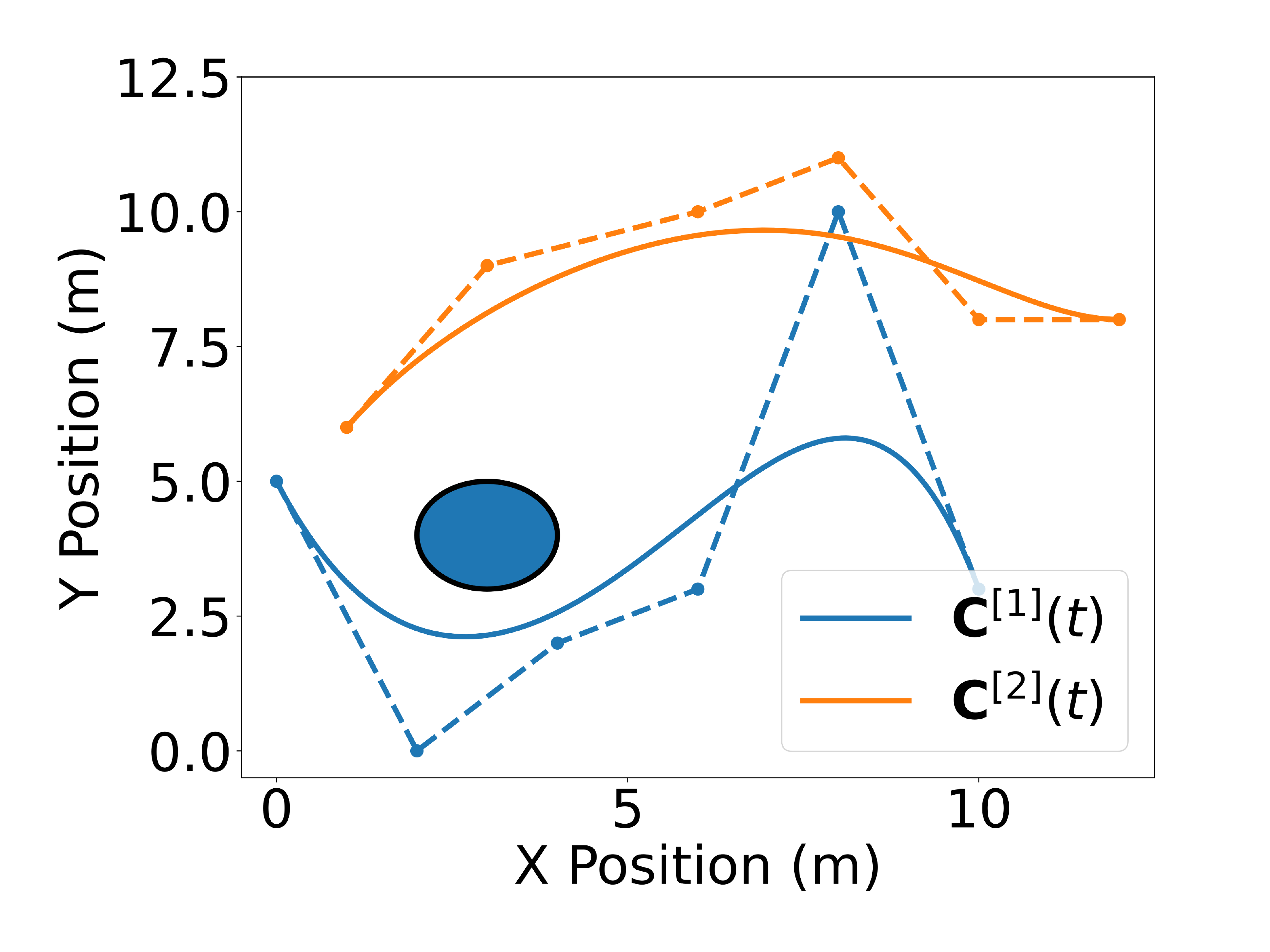}
    \caption{Two Bernstein polynomial trajectories in 2D near a circular obstacle. Trajectory $\mathbf C^{[1]}(t)$ is drawn in blue and trajectory $\mathbf C^{[2]}(t)$ is drawn in orange.}
    \label{fig:21init}
\end{subfigure}
\hfill
\begin{subfigure}[t]{0.49\textwidth}
    \centering
    \includegraphics[scale=0.2]{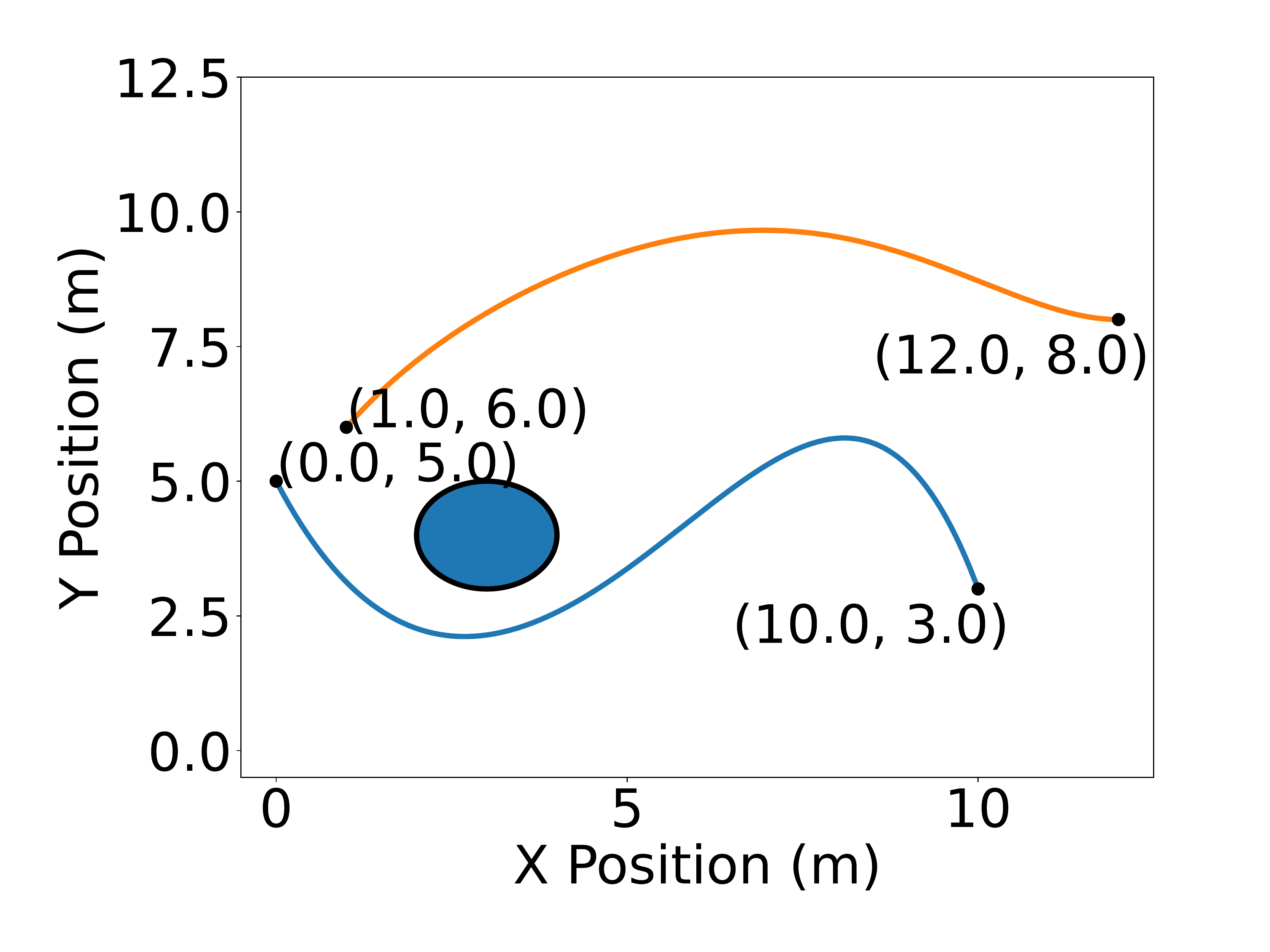}
    \caption{Trajectories $\mathbf C^{[1]}(t)$ and $\mathbf C^{[2]}(t)$ with their endpoints highlighted in 2D.}
    \label{fig:21endpts}
\end{subfigure}
\begin{subfigure}[t]{0.49\textwidth}
    \centering
    \includegraphics[scale=0.2]{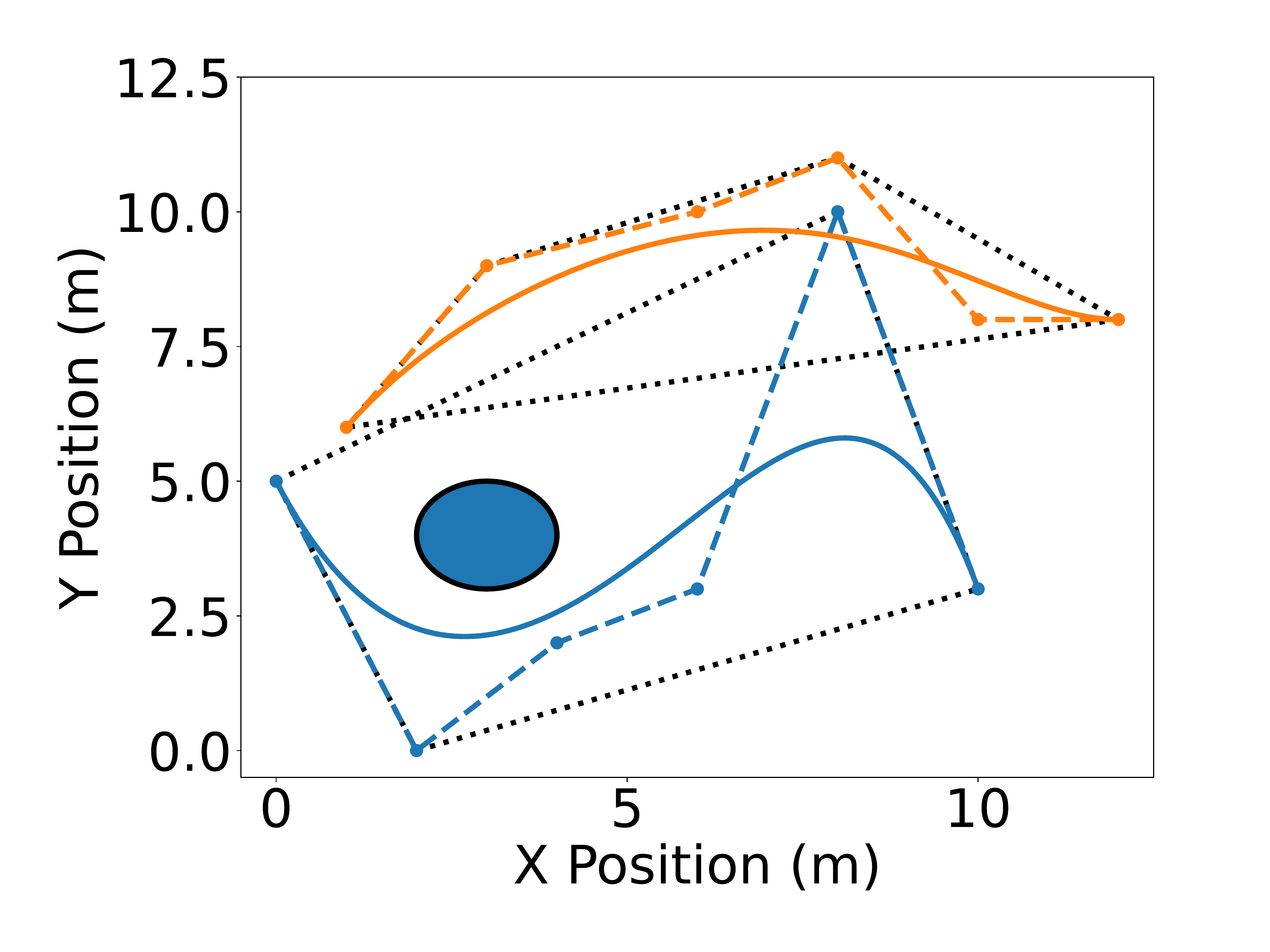}
    \caption{Convex hulls drawn as black dotted lines around the Bernstein coefficients of trajectories $\mathbf C^{[1]}(t)$ and $\mathbf C^{[2]}(t)$.}
    \label{fig:21convexhull}
\end{subfigure}
\hfill
\begin{subfigure}[t]{0.49\textwidth}
    \centering
    \includegraphics[scale=0.2]{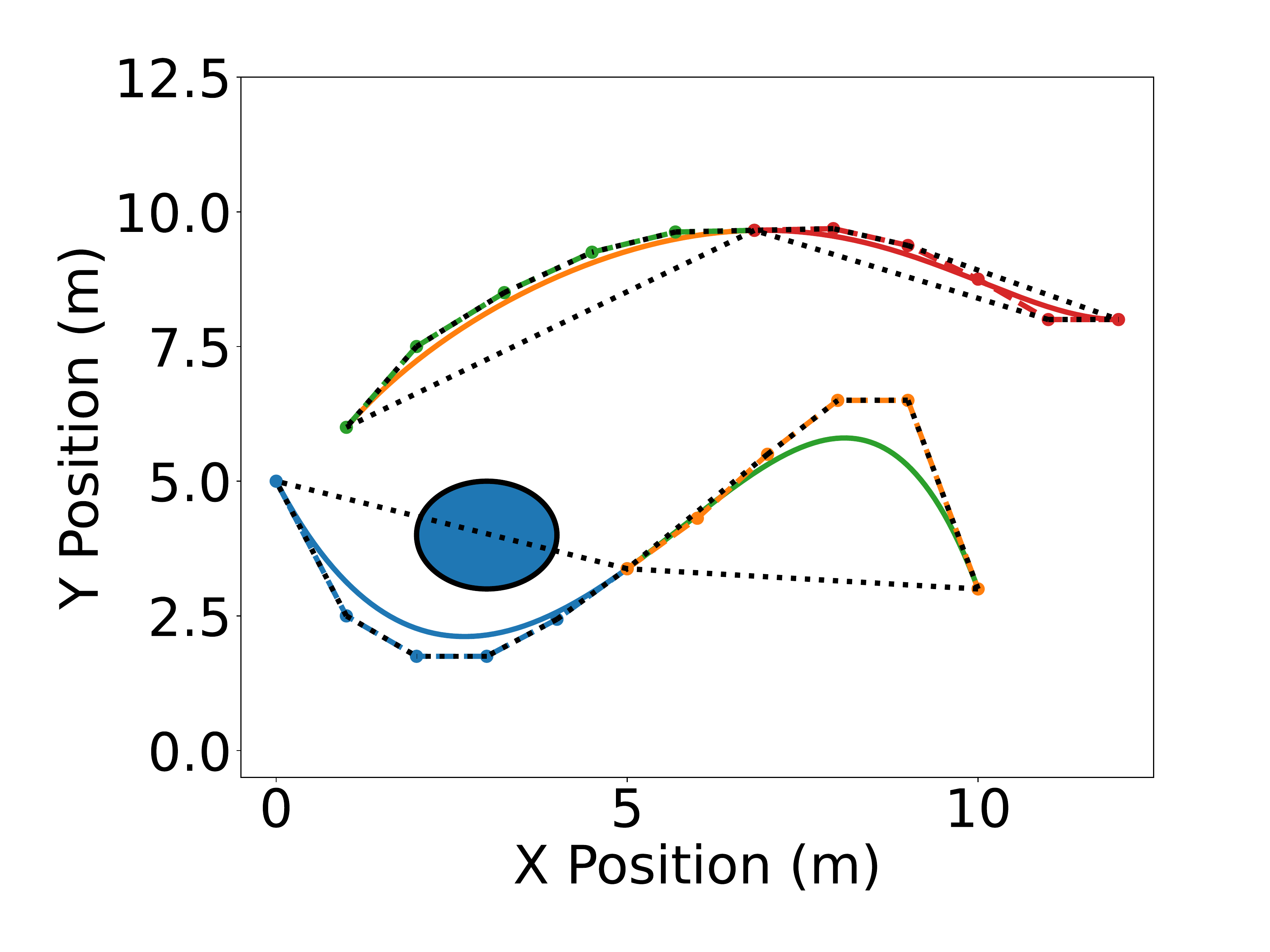}
    \caption{Trajectories $\mathbf C^{[1]}(t)$ and $\mathbf C^{[2]}(t)$ split at $t_{\textit{div}} = 15 s$. Convex hulls are drawn around the Bernstein coefficients of the new split trajectories.}
    \label{fig:21splitconvexhull}
\end{subfigure}
\begin{subfigure}[t]{0.49\textwidth}
    \centering
    \includegraphics[scale=0.2]{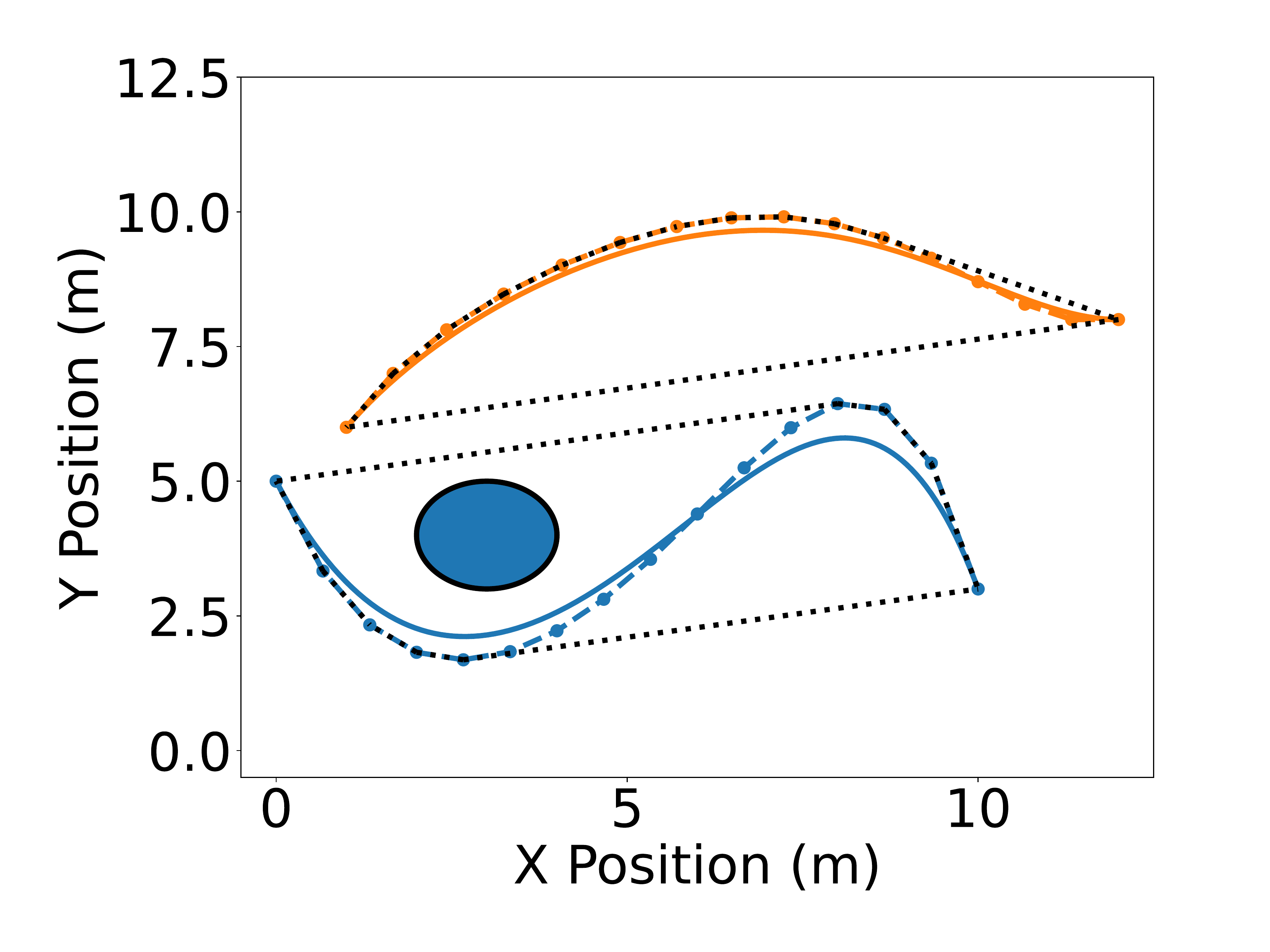}
    \caption{Convex hull drawn around the elevated Bernstein coefficients of trajectories $\mathbf C^{[1]}(t)$ and $\mathbf C^{[2]}(t)$.}
    \label{fig:21elevatedconvexhull}
\end{subfigure}
\hfill
\begin{subfigure}[t]{0.49\textwidth}
    \centering
    \includegraphics[scale=0.2]{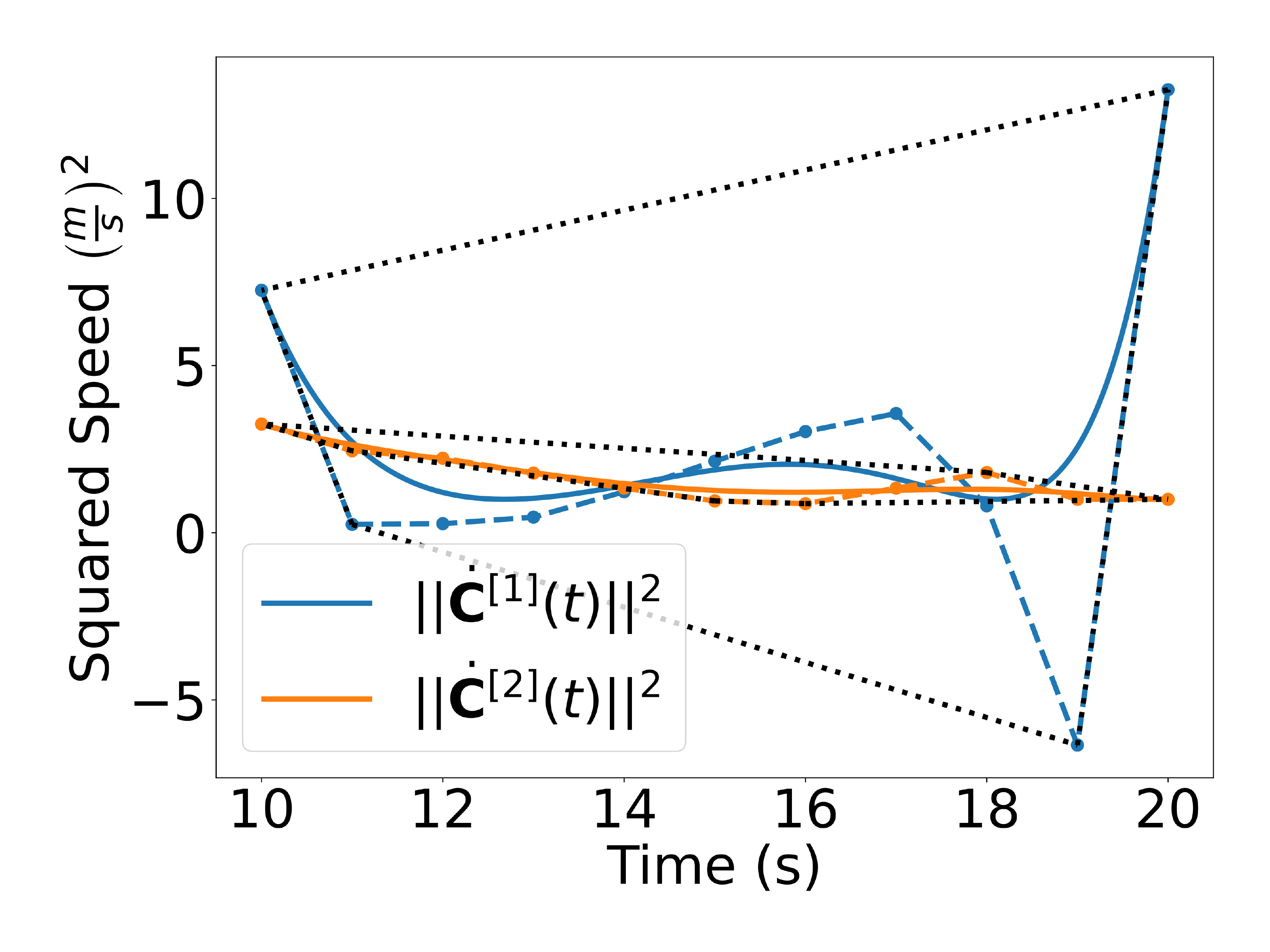} 
    \caption{Squared speed of the trajectories $\mathbf C^{[1]}(t)$ and $\mathbf C^{[2]}(t)$. A convex hull is drawn around the Bernstein coefficients. Note that even though the coefficients may be negative, the actual curve is not.}
    \label{fig:21speedsquared}
\end{subfigure}
\begin{subfigure}[t]{0.49\textwidth}
    \centering
    \includegraphics[scale=0.2]{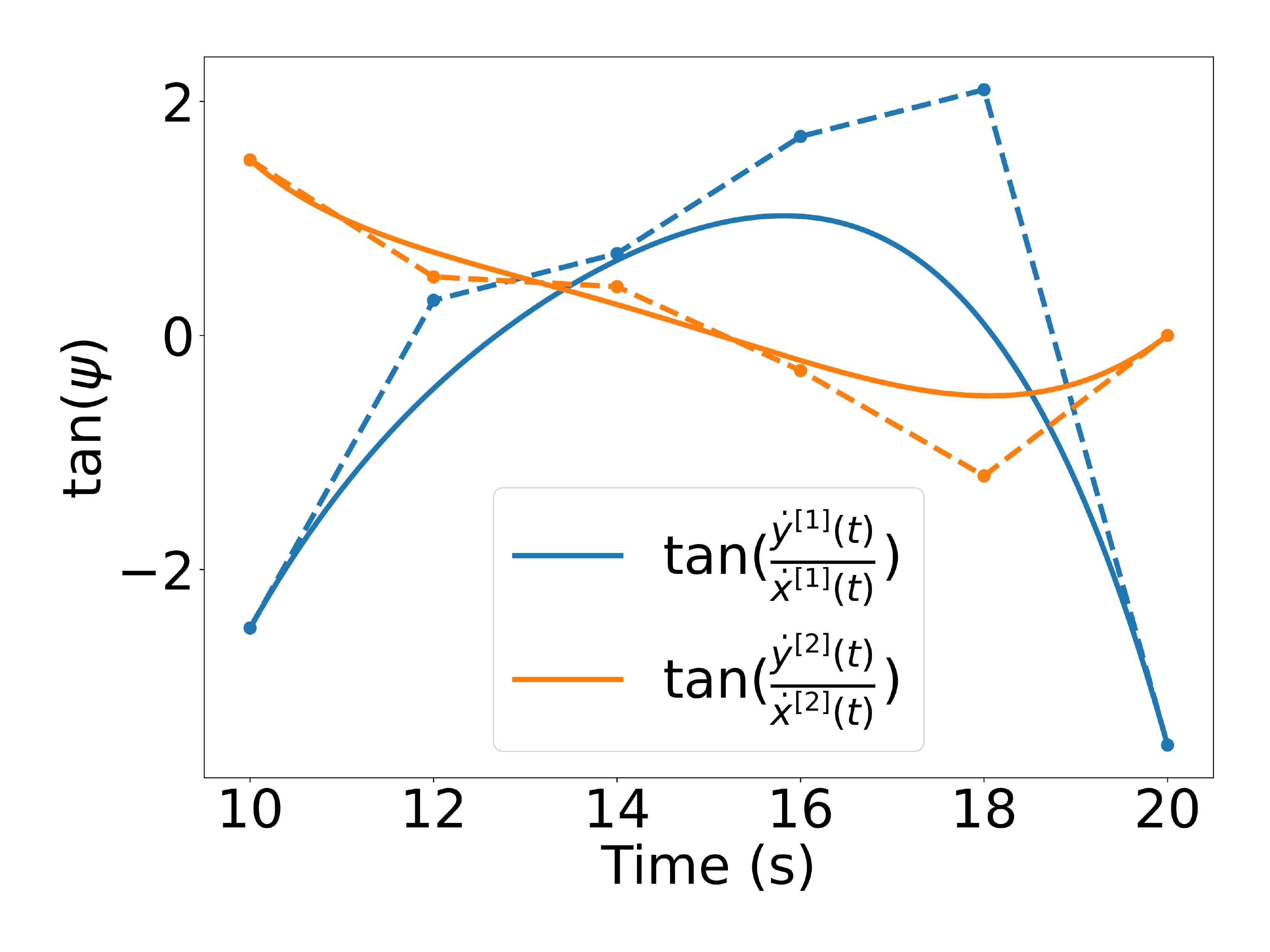}
    \caption{Tangent of the heading angle of trajectories $\mathbf C^{[1]}(t)$ and $\mathbf C^{[2]}(t)$.}
    \label{fig:21headingangle}
\end{subfigure}
\hfill
\begin{subfigure}[t]{0.49\textwidth}
    \centering
    \includegraphics[scale=0.2]{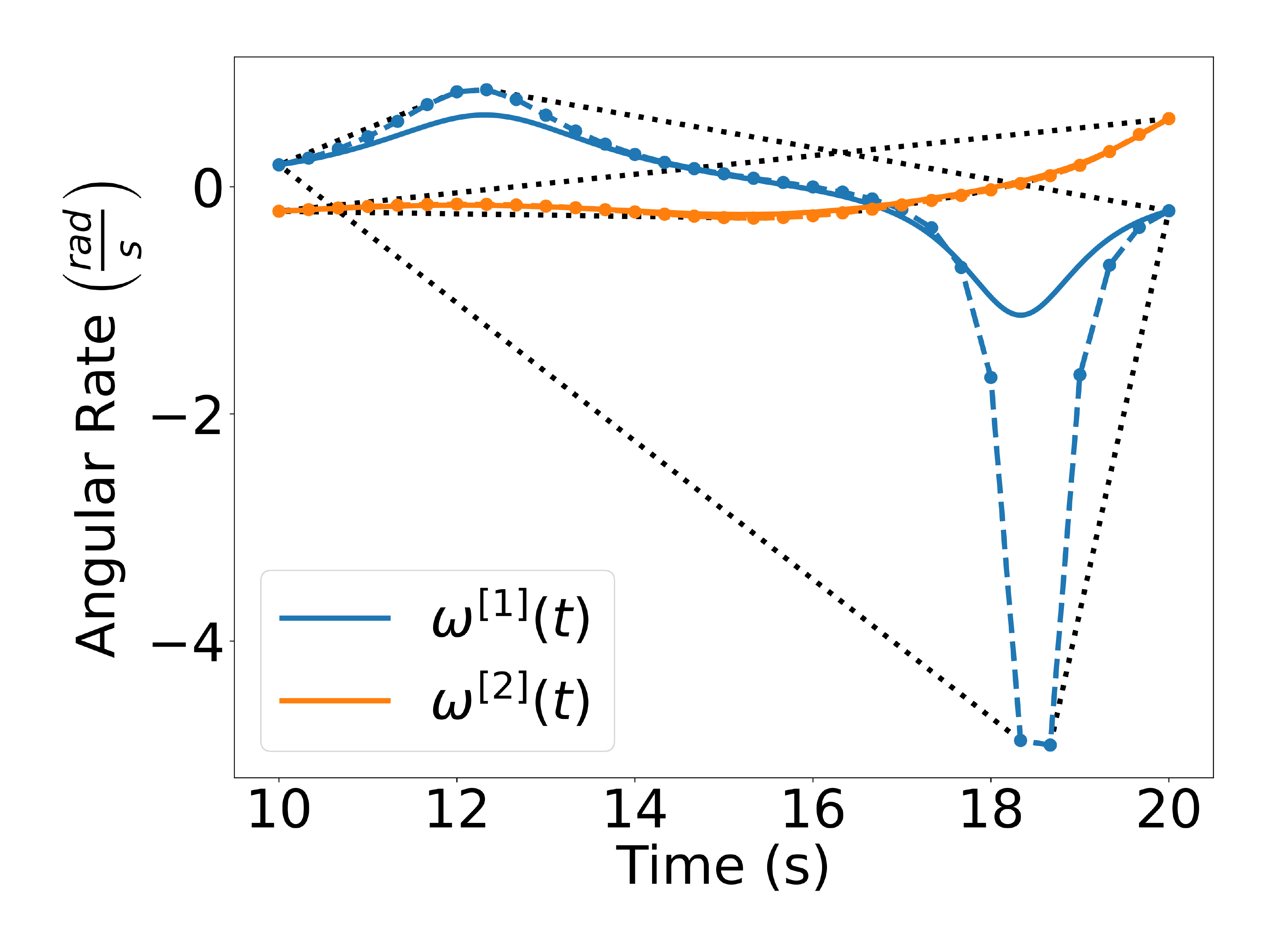}
    \caption{Angular rates of trajectories $\mathbf C^{[1]}(t)$ and $\mathbf C^{[2]}(t)$. Note that the angular rates are rational Bernstein polynomials.}
    \label{fig:21angrate}
\end{subfigure}
\caption{Several illustrative examples of the properties of Bernstein polynomials being applied to 2D trajectories.}
\label{fig:21figs}
\end{figure*}

Figure \ref{fig:21figs} contains several examples of 2-dimensional trajectories. Two trajectories are plotted in Figure \ref{fig:21init} along with an obstacle. The trajectories $\mathbf C^{[1]}(t)$ and $\mathbf C^{[2]}(t)$ are defined as in \eqref{eq:BezCurve} with $t_0 = 10 s$ and $t_f = 20 s$. The Bernstein coefficients of trajectory $\mathbf C^{[1]}(t)$ are
$$
\mathbf{P}_5^{[1]} =
\begin{bmatrix}
0 & 2 & 4 & 6 & 8 & 10 \\
5 & 0 & 2 & 3 & 10 & 3
\end{bmatrix}.
$$
The Bernstein coefficients of trajectory $\mathbf C^{[2]}(t)$ are
$$
\mathbf{P}_5^{[2]} =
\begin{bmatrix}
1 & 3 & 6 & 8 & 10 & 12 \\
6 & 9 & 10 & 11 & 8 & 8
\end{bmatrix}.
$$
The circular obstacle has a radius of $1$ and is centered at point $[3, 4]^\intercal$. Figure \ref{fig:21endpts} highlights the endpoints property (Property \ref{prop:endpts}). Note that the trajectory $\mathbf C^{[1]}(t)$ passes through its first and last Bernstein coefficients which are $\mathbf{P}_{0,5}^{[1]}=[0, 5]^\intercal$ and $[10, 3]^\intercal$, respectively. Likewise, the trajectory $\mathbf C^{[2]}(t)$ passes through its first and last Bernstein coefficients which are $[1, 6]^\intercal$ and $[12, 8]^\intercal$, respectively. The convex hull property (Property \ref{prop:convexhull}) is illustrated in Figure \ref{fig:21convexhull}.

Useful operations can be efficiently performed on Bernstein polynomials by only manipulating the coefficients. The de Casteljau algorithm (Property \ref{prop:decast}) allows one to split a Bernstein polynomial into two separate polynomials. This is shown in Figure \ref{fig:21splitconvexhull} where trajectories $\mathbf C^{[1]}(t)$ and $\mathbf C^{[2]}(t)$ are split at $t_{\textit{div}} = 15 s$. Degree elevation (Property \ref{prop:elevation}) is performed on both trajectories in Figure \ref{fig:21elevatedconvexhull}. Note that in both Figure \ref{fig:21splitconvexhull} and \ref{fig:21elevatedconvexhull} the convex hulls are more accurate than the conservative convex hulls in Figure \ref{fig:21convexhull}. This idea will be expanded upon in the next section.

Bernstein polynomials can also be used to extract useful dynamics and control information. In Figure \ref{fig:21speedsquared} Bernstein polynomials representing the squared speed of trajectories $\mathbf C^{[1]}(t) = [x^{[1]}(t), y^{[1]}(t)]^\intercal$ and $\mathbf C^{[2]}(t) = [x^{[2]}(t), y^{[2]}(t)]^\intercal$ is shown along with their corresponding coefficients and convex hulls. The squared speed is computed using the derivative and arithmetic operation properties (Properties \ref{prop:derivatives} and \ref{prop:arithmetic}) as follows
\begin{equation*}
    (v^{[1]}(t))^2 = (\dot x^{[1]}(t))^2 + (\dot y^{[1]}(t))^2
\end{equation*}
Note that the squared speed of a trajectory described by a Bernstein polynomial is also a Bernstein polynomial.

Figure \ref{fig:21headingangle} shows the tangent of the heading angle of trajectories $\mathbf C^{[1]}(t)$ and $\mathbf C^{[2]}(t)$. For example, letting $\mathbf C^{[1]}(t) = [x^{[1]}(t) \, , \, y^{[1]}(t)]^\top$, the heading angle $\psi(t)$ of a trajectory can be found by
\begin{equation} \label{eq:headingangle}
    \psi^{[1]}(t) = \tan^{-1} \left( \frac{\dot y^{[1]}(t)}{\dot x^{[1]}(t)} \right).
\end{equation}
Since the inverse tangent of a Bernstein polynomial is not a Bernstein polynomial, we take the tangent of both sides of the equation and represent the tangent of the heading angle as a rational Bernstein polynomial (see Figure \ref{fig:21headingangle}). 

To determine the angular rate, we can take the derivative of the heading angle,
\begin{equation} \label{eq:angrate}
    \omega^{[1]}(t) = \dot \psi^{[1]}(t) = \frac{\dot x^{[1]}(t) \ddot y^{[1]}(t) - \ddot x^{[1]}(t) \dot y^{[1]}(t)}{(\dot x^{[1]}(t))^2 + (\dot y^{[1]}(t))^2}.
\end{equation}
Since the angular rate can be determined using Properties \ref{prop:derivatives} and \ref{prop:arithmetic}, it can be represented as a rational Bernstein polynomial. The angular rates of trajectories $\mathbf C^{[1]}(t)$ and $\mathbf C^{[2]}(t)$ are shown in Figure \ref{fig:21angrate}.

Finally, the Bernstein polynomial representing the squared distance between two trajectories at every point in time can be found by
\begin{equation}
    \begin{aligned}
        d^2(t) &= (x^{[2]}(t) - x^{[1]}(t))^2 + (y^{[2]}(t) - y^{[1]}(t))^2, \\
        &\forall t \in [t_0, t_f]
    \end{aligned}
\end{equation}
where $d^2(t)$ is the squared distance between two trajectories whose positions are defined by $[x^{[1]}(t), y^{[1]}(t)]^\intercal$ and $[x^{[2]}(t), y^{[2]}(t)]^\intercal$.
The position of a static obstacle $\mathbf{Obs}(t)$ can be represented as a Bernstein polynomial whose coefficients are all identical and set to the position of the obstacle, i.e.
$$
\mathbf P^{[Obs]} =
\begin{bmatrix}
x^{[Obs]} & \cdots & x^{[Obs]} \\
y^{[Obs]} & \cdots & y^{[Obs]}
\end{bmatrix}.
$$
The degree of the Bernstein polynomial representing the position of the obstacle is equal to that of the order of the Bernstein polynomials representing the trajectories.
In order to represent the speed and distance as Bernstein polynomials, it is necessary to use their squared values because the square root of a Bernstein polynomial is not a Bernstein polynomial.

\subsection{Numerical Examples in 3D}

\begin{figure*}
\centering
\begin{subfigure}[t]{0.49\textwidth}
    \centering
    \includegraphics[scale=0.2]{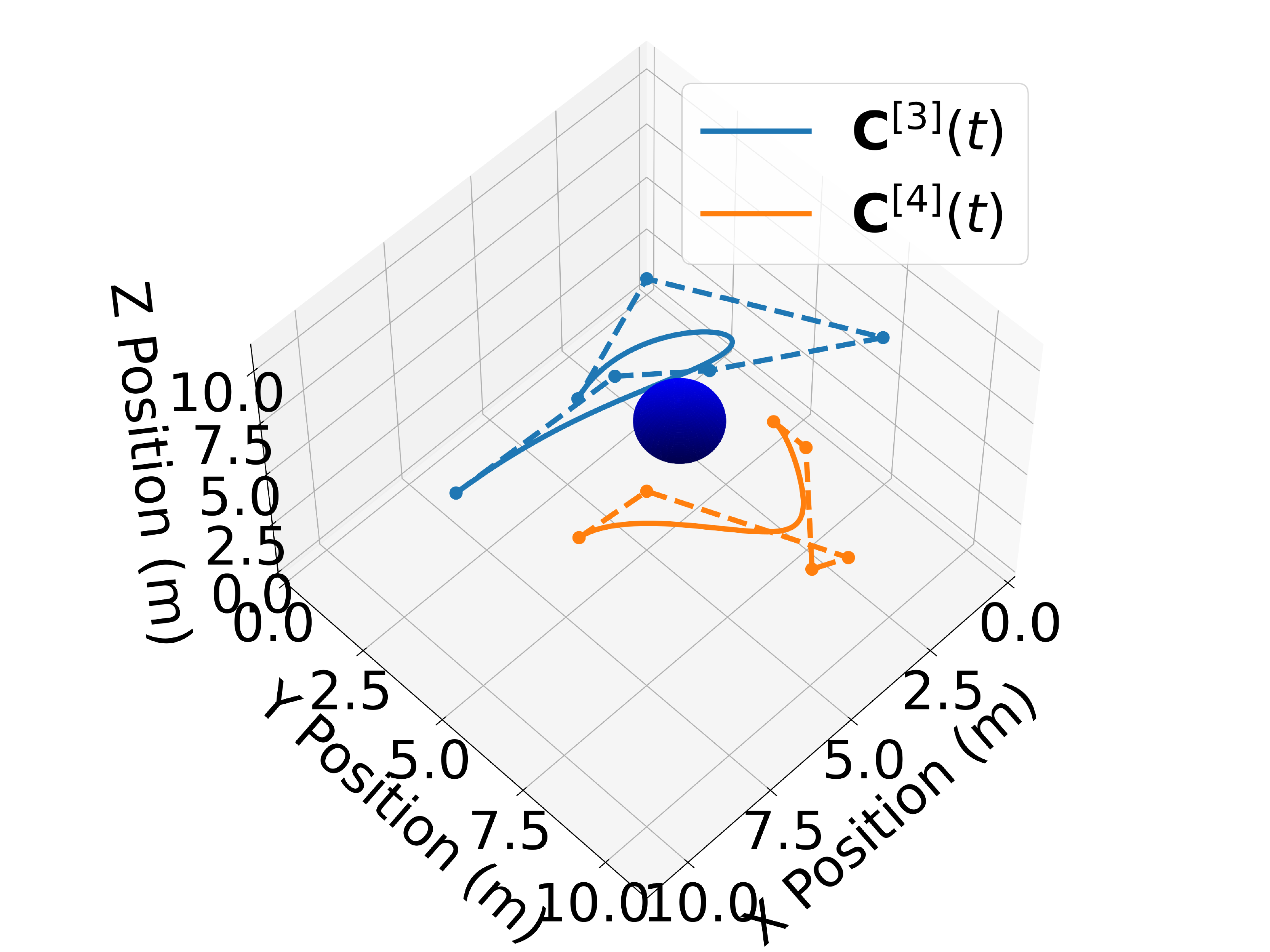}
    \caption{Two 3D Bernstein polynomial trajectories near a spherical obstacle. Trajectory $\mathbf C^{[3]}(t)$ is drawn in blue and trajectory $\mathbf C^{[4]}(t)$ is drawn in orange.}
    \label{fig:22init}
\end{subfigure}
\hfill
\begin{subfigure}[t]{0.49\textwidth}
    \centering
    \includegraphics[scale=0.2]{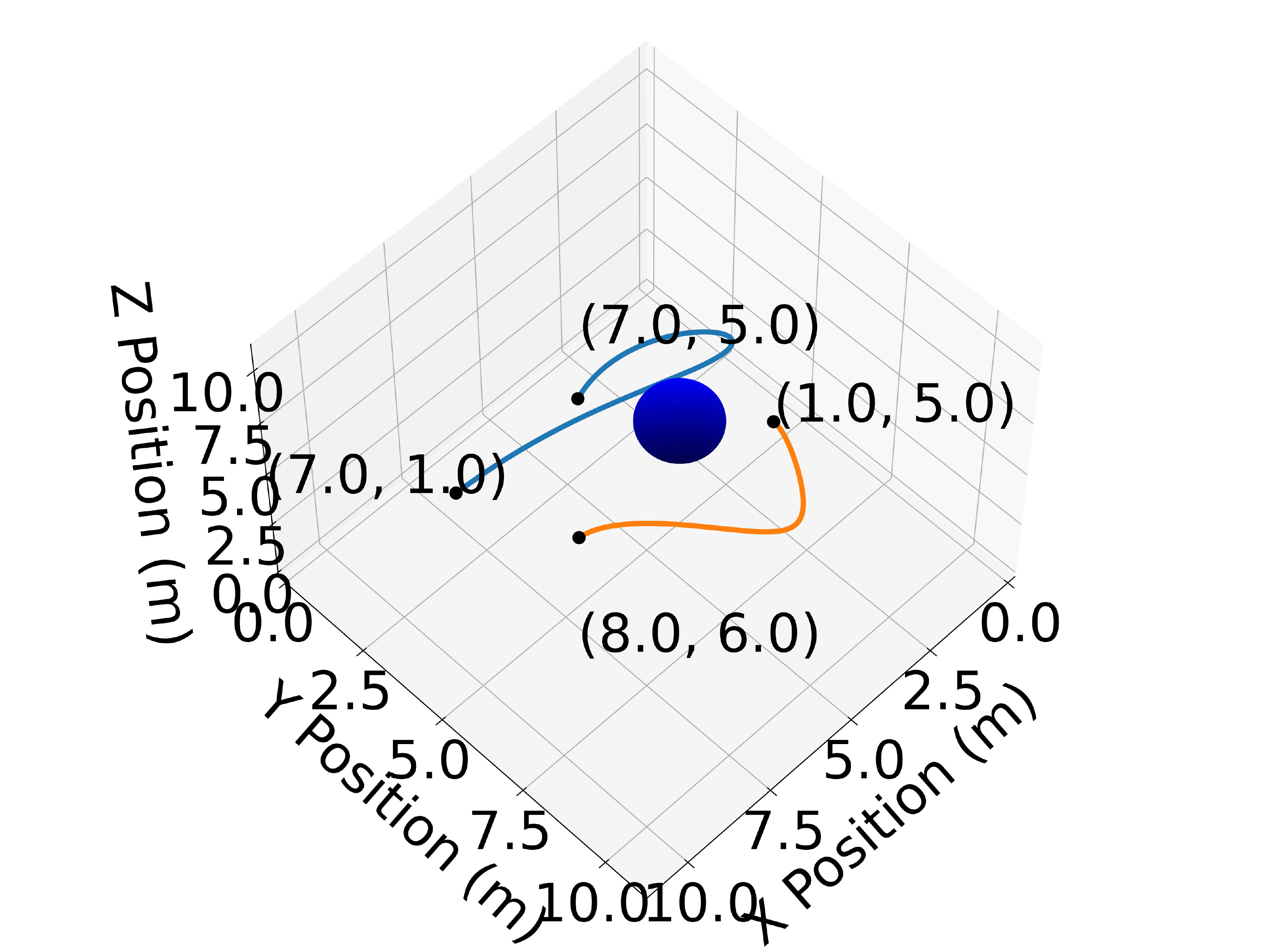}
    \caption{3D trajectories $\mathbf C^{[3]}(t)$ and $\mathbf C^{[4]}(t)$ with their endpoints highlighted.}
    \label{fig:22endpts}
\end{subfigure}
\begin{subfigure}[t]{0.49\textwidth}
    \centering
    \includegraphics[scale=0.2]{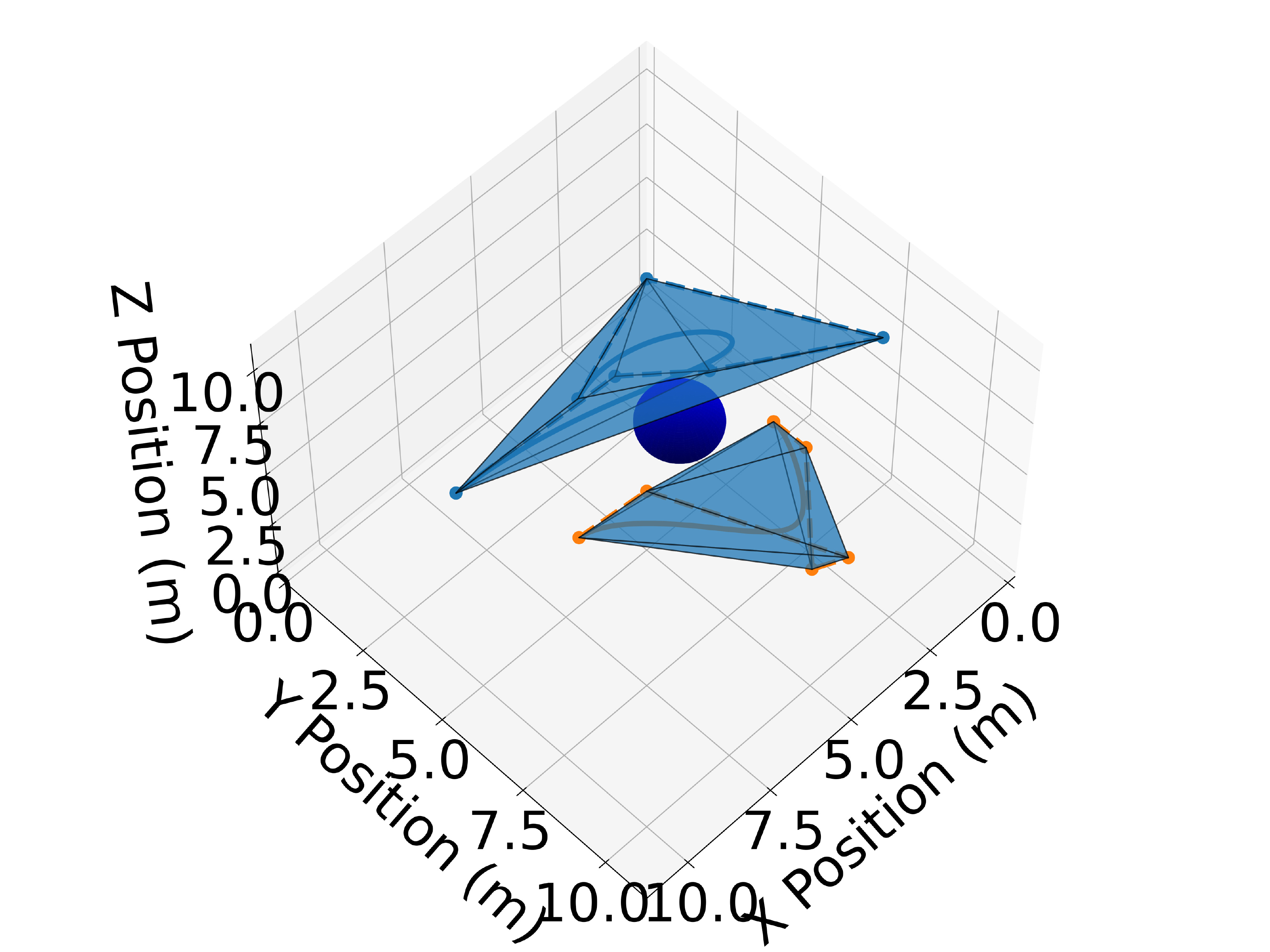}
    \caption{3D convex hulls drawn as transparent blue surfaces around the Bernstein coefficients of trajectories $\mathbf C^{[3]}(t)$ and $\mathbf C^{[4]}(t)$.}
    \label{fig:22convexhull}
\end{subfigure}
\hfill
\begin{subfigure}[t]{0.49\textwidth}
    \centering
    \includegraphics[scale=0.2]{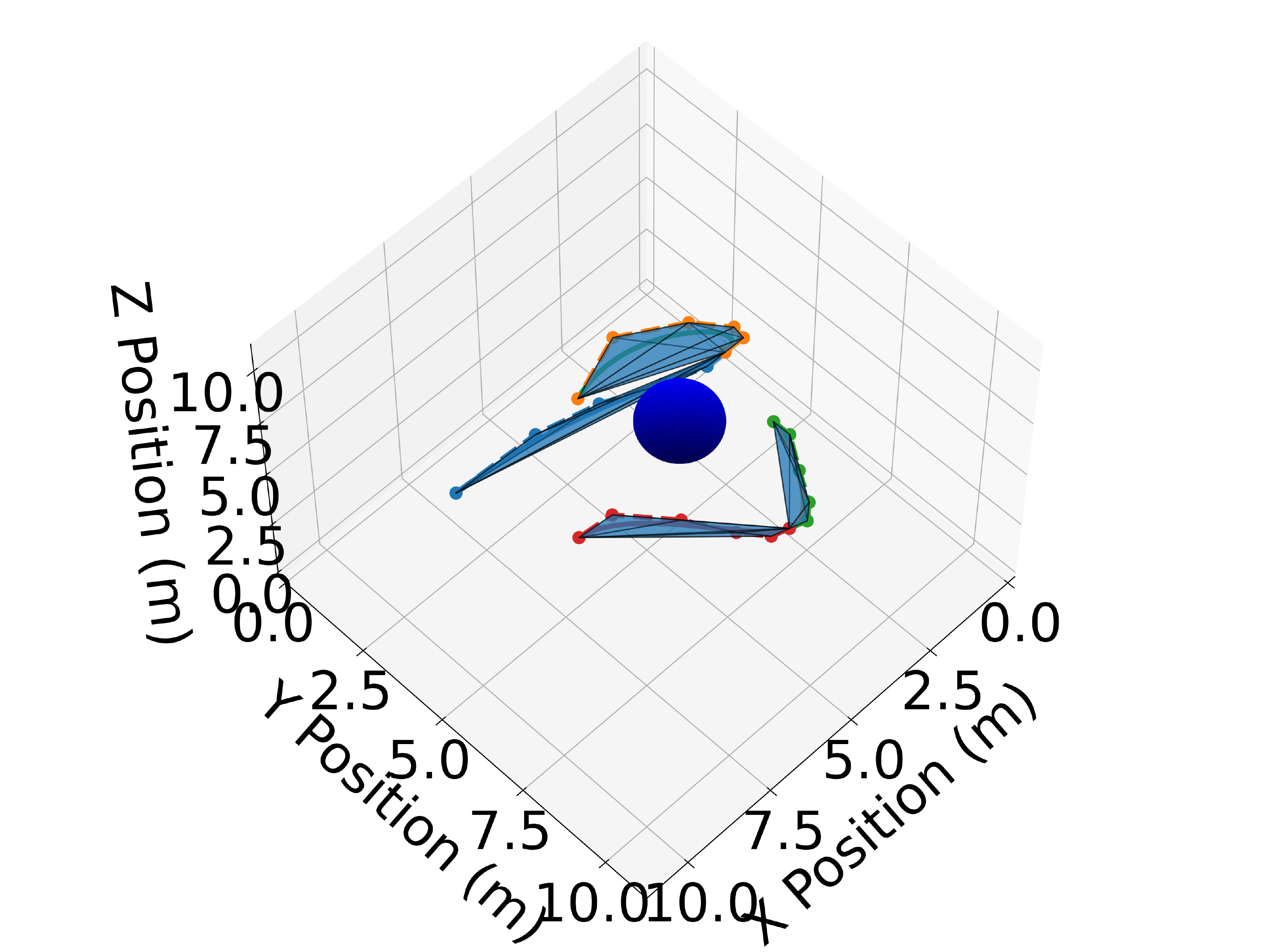}
    \caption{Trajectories $\mathbf C^{[3]}(t)$ and $\mathbf C^{[4]}(t)$ split at $t_{\textit{div}} = 15 s$. Convex hulls are drawn around the Bernstein coefficients of the new split trajectories.}
    \label{fig:22splitconvexhull}
\end{subfigure}
\begin{subfigure}[t]{0.49\textwidth}
    \centering
    \includegraphics[scale=0.2]{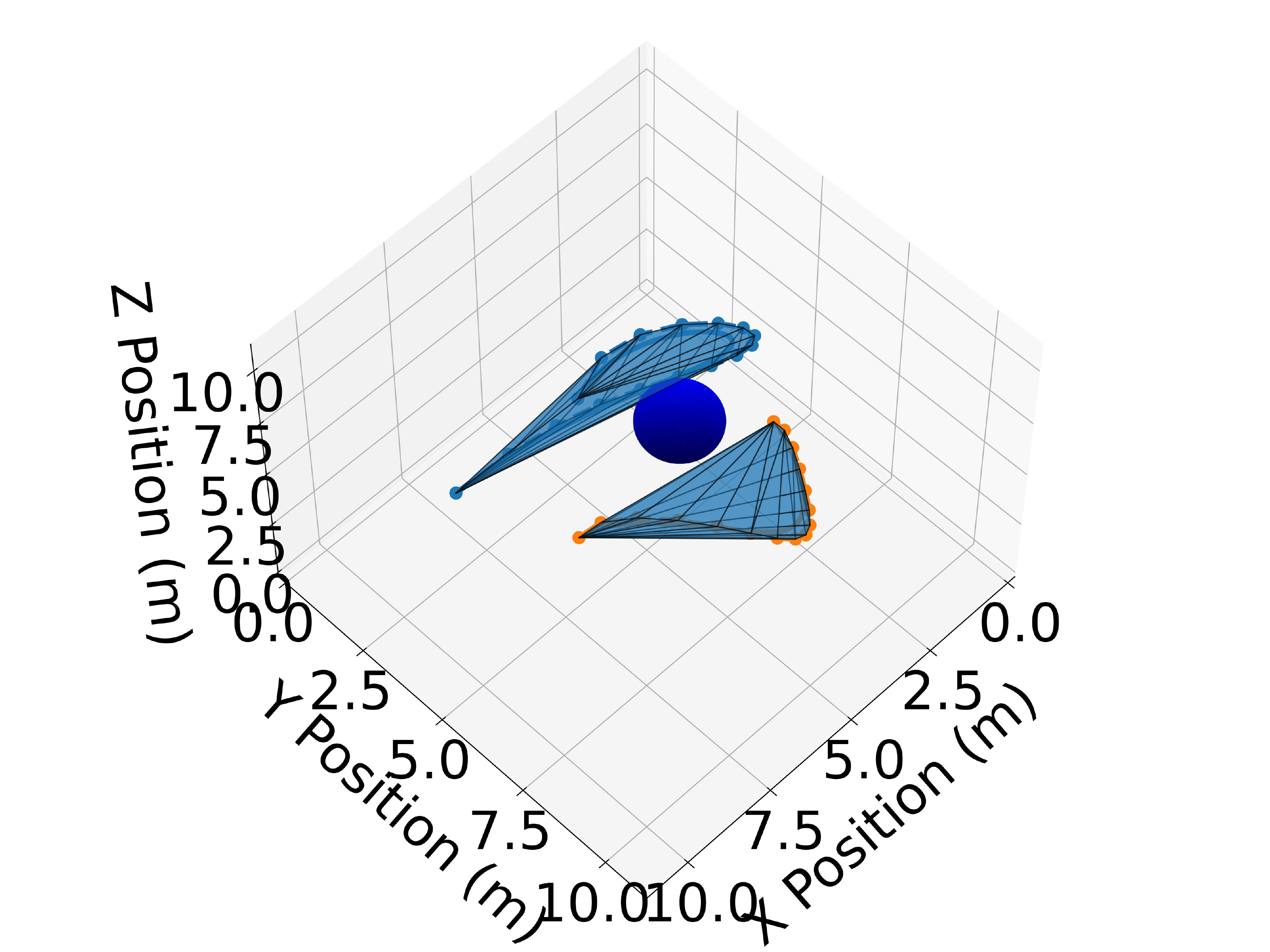}
    \caption{Convex hull drawn around the elevated Bernstein coefficients of trajectories $\mathbf C^{[3]}(t)$ and $\mathbf C^{[4]}(t)$.}
    \label{fig:22elevatedconvexhull}
\end{subfigure}
\hfill
\begin{subfigure}[t]{0.49\textwidth}
    \centering
    \includegraphics[scale=0.2]{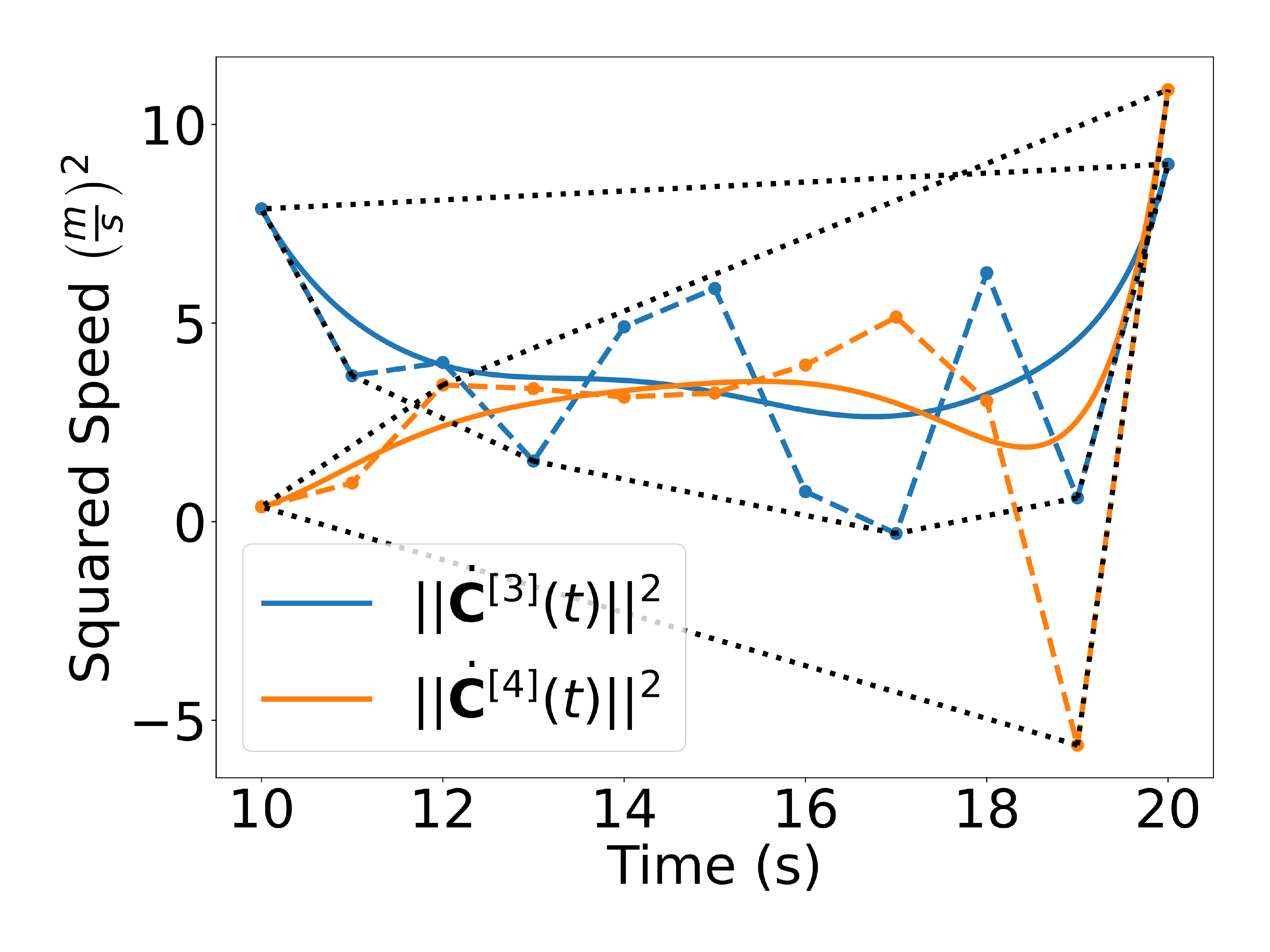}
    \caption{Squared speed of the trajectories $\mathbf C^{[3]}(t)$ and $\mathbf C^{[4]}(t)$. A convex hull is drawn around the Bernstein coefficients. Note that even though the coefficients may be negative, the actual curve is not.}
    \label{fig:22speedsquared}
\end{subfigure}
\begin{subfigure}[t]{0.49\textwidth}
    \centering
    \includegraphics[scale=0.2]{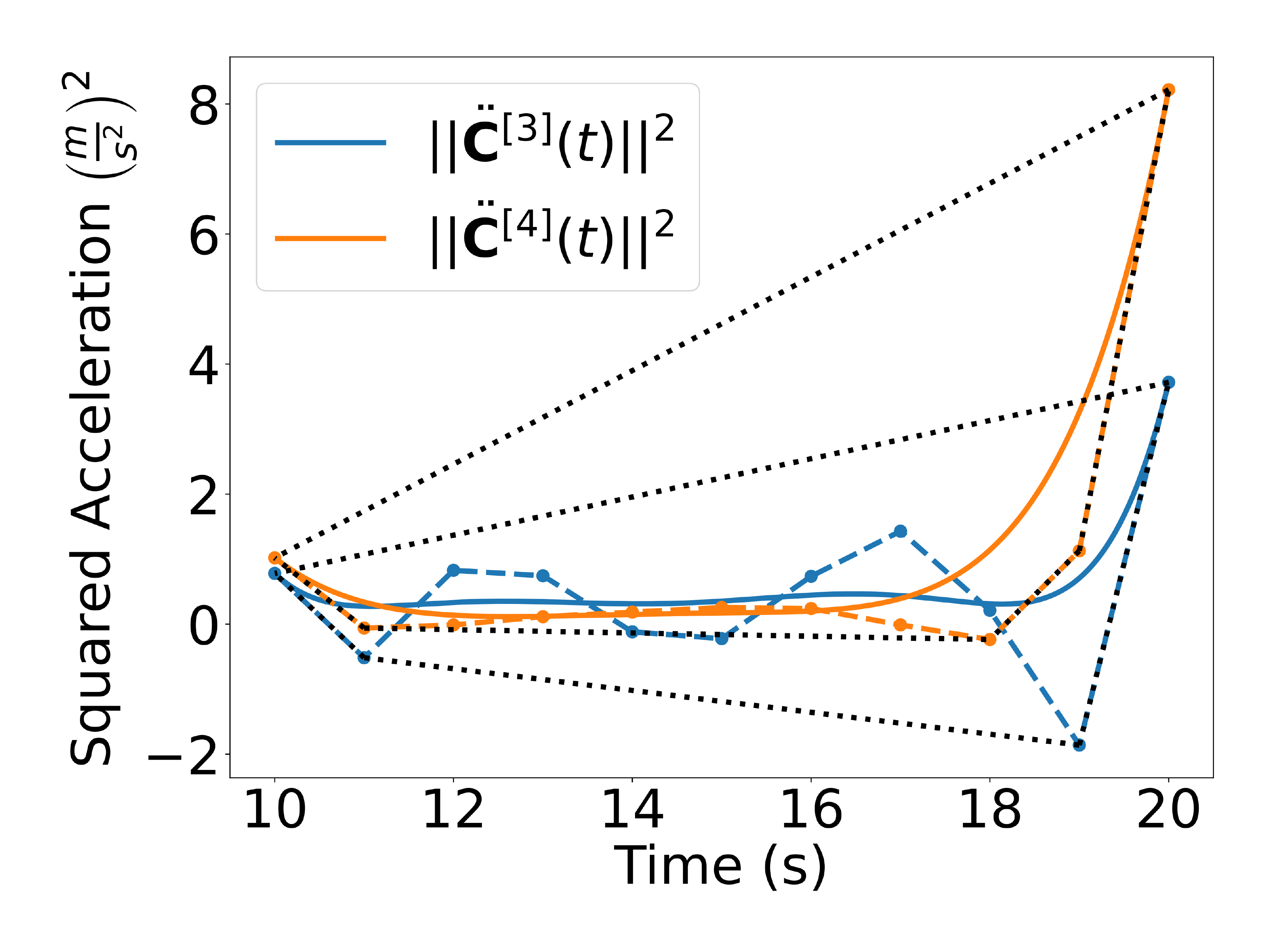}
    \caption{Squared acceleration of the trajectories $\mathbf C^{[3]}(t)$ and $\mathbf C^{[4]}(t)$ with corresponding convex hulls.}
    \label{fig:22accelsquared}
\end{subfigure}
\hfill
\begin{subfigure}[t]{0.49\textwidth}
    \centering
    \includegraphics[scale=0.2]{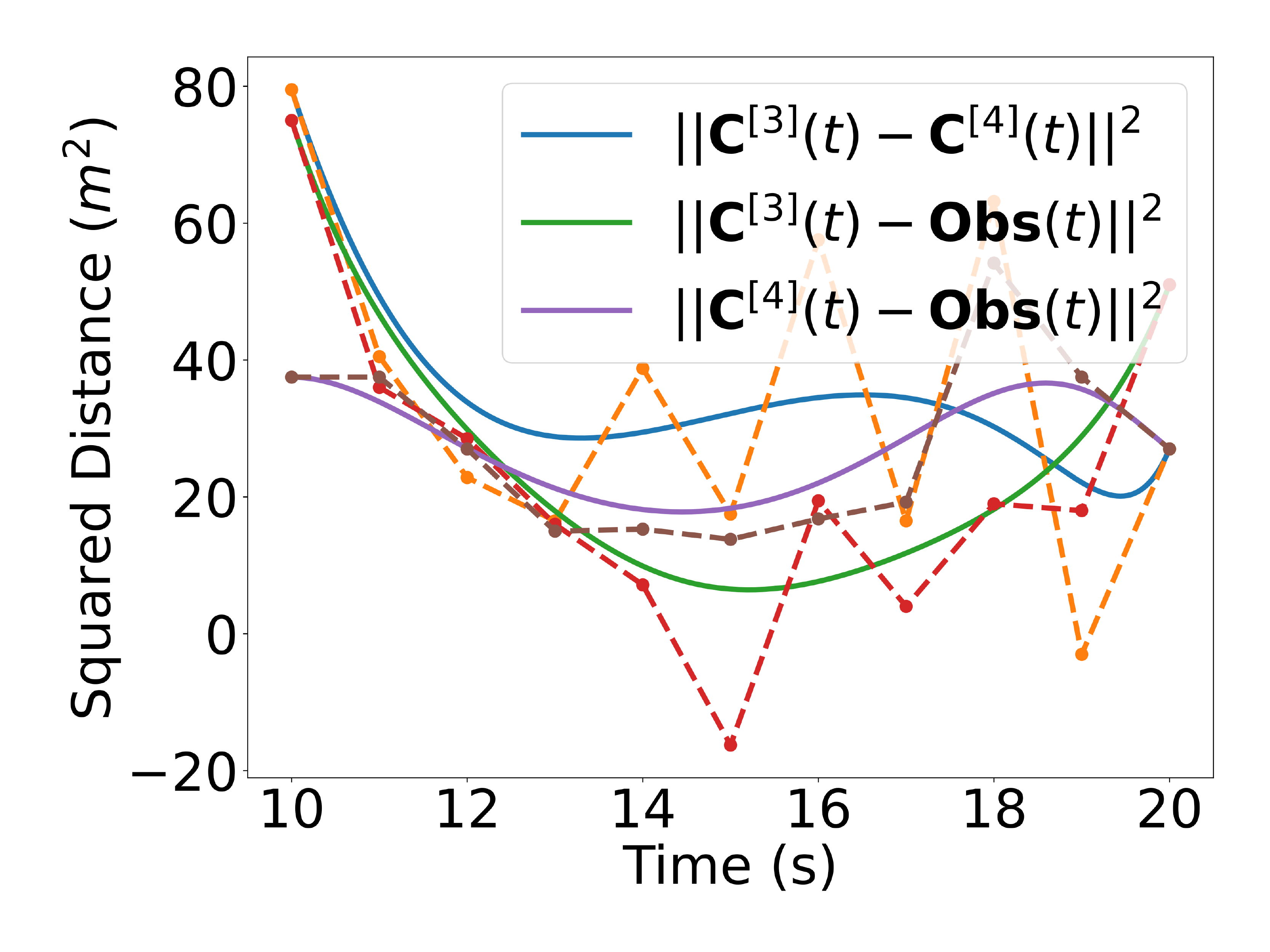}
    \caption{Squared distances between the trajectories and then center of the spherical obstacle.}
    \label{fig:22distsquared}
\end{subfigure}
\caption{Several illustrative examples of the properties of Bernstein polynomials being applied to 3D trajectories.}
\label{fig:22figs}
\end{figure*}

Next, we introduce two 3D Bernstein polynomials with $t_0 = 10 s$ and $t_f = 20 s$ and illustrate their properties in Figure \ref{fig:22figs}. The Bernstein coefficients of trajectory $\mathbf C^{[3]} (t)$ are
$$
\mathbf{P}_5^{[3]} =
\begin{bmatrix}
7 & 3 & 1 & 1 & 3 & 7 \\
1 & 2 & 3 & 8 & 3 & 5 \\
0 & 2 & 1 & 9 & 8 & 10
\end{bmatrix},
$$
and the Bernstein coefficients of trajectory $\mathbf C^{[4]} (t)$ are
$$
\mathbf{P}_5^{[4]} =
\begin{bmatrix}
1 & 1 & 4 & 4 & 8 & 8 \\
5 & 6 & 9 & 10 & 8 & 6 \\
1 & 1 & 3 & 5 & 11 & 6
\end{bmatrix}.
$$
These polynomials are drawn in Figure \ref{fig:22init}.

Similar to the 2D examples, Figures \ref{fig:22endpts}, \ref{fig:22convexhull}, \ref{fig:22splitconvexhull}, and \ref{fig:22elevatedconvexhull} illustrate the end points, convex hull, de Casteljau, and elevation properties, respectively. Figures \ref{fig:22speedsquared} and \ref{fig:22accelsquared} show the squared speed and squared acceleration of trajectories $\mathbf C^{[3]}(t)$ and $\mathbf C^{[4]}(t)$, respectively. These values were computed using the derivative and arithmetic properties. Finally, Figure \ref{fig:22distsquared} shows the squared Euclidean distance between the trajectories and the center of the spherical obstacle at every point in time. The distance was found using the arithmetic properties.

\begin{figure}
    \centering
    \includegraphics[scale=0.2]{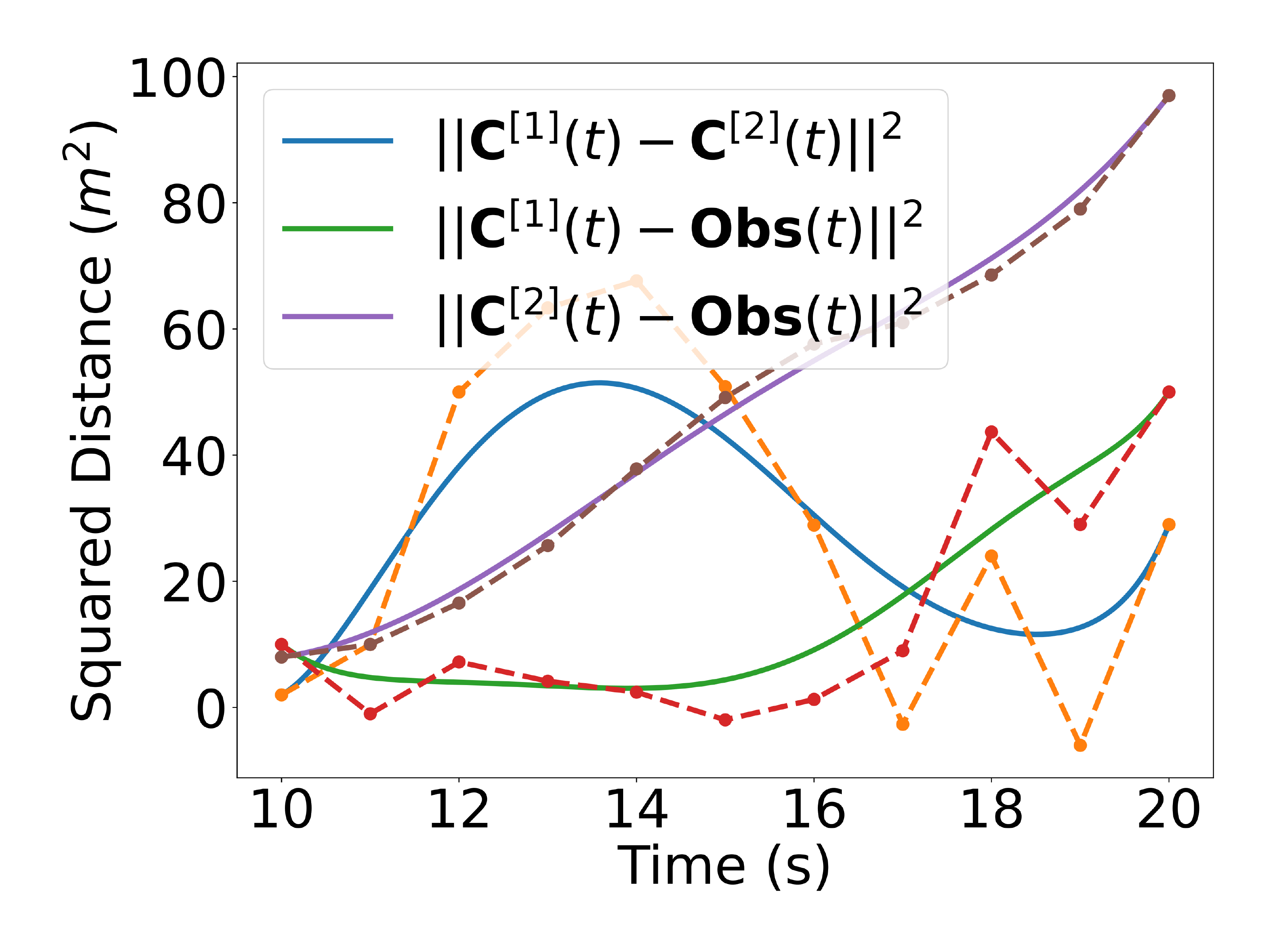}
    \caption{Squared distance between trajectories and the center of the circular obstacle.}
    \label{fig:21squareddist}
\end{figure}

\section{Algorithms and Functions} \label{sec:algorithms}
This section presents algorithms and procedures for Bernstein polynomials that use the properties listed above.

\subsection{Evaluating Bounds}
\label{sec:evaluatingbounds}
\begin{figure}
    \centering
    \includegraphics[scale=0.19, trim=1cm 0 0 0]{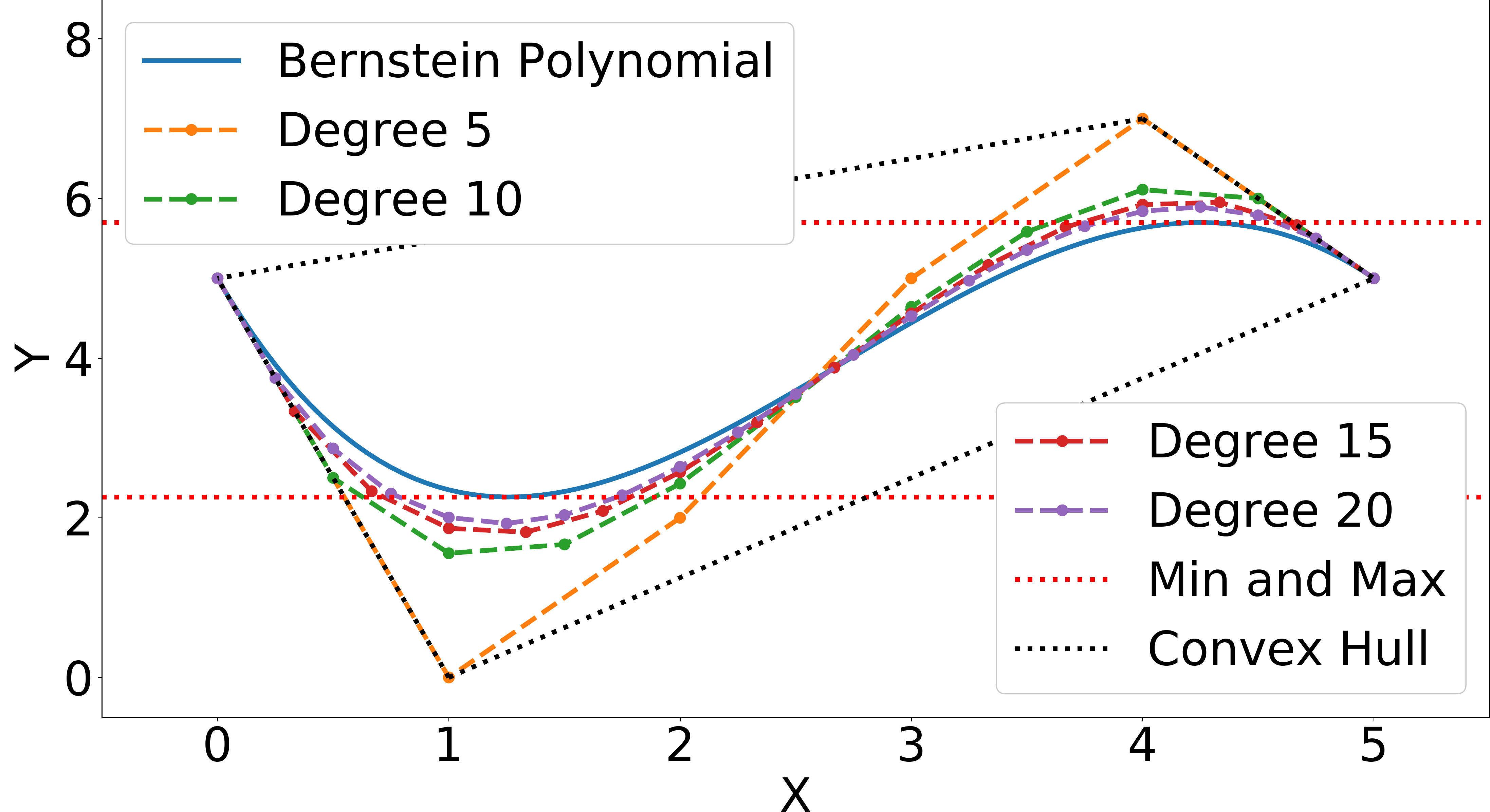}
    \caption{Bounds for Bernstein polynomials. The solid blue line is the Bernstein polynomial, the dashed lines connect the coefficients of each different order, the black dotted line represents the convex hull of the 5th degree Bernstein polynomial, and the red dotted lines represent the actual extrema.}
    \label{fig:elevminmaxhull}
\end{figure}

Using Property \ref{prop:convexhull}, one can quickly determine conservative bounds of the Bernstein polynomial.
For example, given the $1$-dimensional Bernstein polynomial introduced in \eqref{eq:BezCurve1dim}, with coefficients given by
$$
\mathbf{P}_5 =
\begin{bmatrix}
0 & 1 & 2 & 3 & 4 & 5 \\
5 & 0 & 2 & 5 & 7 & 5
\end{bmatrix} \, ,
$$
lower and upper bounds can be derived using \eqref{eq:convexhull}. The Bernstein polynomial (blue line), and the coefficients (orange dots connected with dashes) are illustrated in Figure \ref{fig:elevminmaxhull}. The lower and upper bounds of the Bernstein polynomial are determined by the values of the minimum and maximum coefficients ($0$ and $7$, respectively), while the actual minimum and maximum of the curve are 2.26 and 5.70, respectively (see horizontal red dotted lines).
While the bounds are conservative, the Bernstein polynomial can be degree elevated. As discussed earlier (see Property \ref{prop:elevation} and \eqref{eq:degelevconv}) the coefficients of a degree elevated Bernstein polynomial converge towards the curve. Thus, they can be used to derive tighter bounds.
Figure \ref{fig:elevminmaxhull} shows the same Bernstein polynomial elevated to orders 10, 15, and 20. The new estimated minimum and maximum values are 1.93 and 5.89, respectively, for the polynomial elevated to order 20.
Degree elevation is performed by multiplying the coefficients of a Bernstein polynomial with the elevation matrix given by \eqref{eq:degelevmatrix}. A database of elevation matrices can be pre-computed off-line to produce tight bounds at a low computational cost.

\subsection{Evaluating Extrema}
\label{sec:evaluatingextrema}
The extrema of a $1$-dimensional Bernstein polynomial are calculated using an iterative procedure similar to the one proposed in \cite{chang2011computation}. Algorithm \ref{algorithm:extrema}, which is used to compute the maximum of a Bernstein polynomial, is described with the understanding that a similar procedure can evaluate the minimum value. The input to Algorithm \ref{algorithm:extrema} is a set containing the coefficients of the Bernstein polynomial, $\mathcal{P} = \{\mathbf{P}_n\}$, $\mathbf{P}_n=[P_{0,n},\ldots,P_{n,n}]$, an arbitrarily large \textit{negative} global maximum value $\alpha$, and a desired tolerance $\epsilon$.

The algorithm first determines a lower bound on the maximum value. This is done by picking the larger of the two end points since, by the end point property (Property \ref{prop:endpts}), the curve is guaranteed to pass through the two points. Next, the upper bound $\mathcal P_{ub}$ is computed by simply finding the largest Bernstein coefficient. The corresponding index of the upper bound $i_{ub}$ is also computed for later use. The first \textit{if} statement replaces the global maximum with the lower bound if the lower bound is larger. This means that the choice of the global maximum value $\alpha$ should be lower than all other expected values otherwise the user will risk returning an incorrect value.

The second \textit{if} statement prunes the recursive tree. If $\alpha$ is greater than the upper bound of the current iteration, there is no reason to continue splitting the current curve since, by the convex hull property (Property \ref{prop:convexhull}), no larger value will be found on that curve.

The third \textit{if} statement checks to see if the upper and lower bounds are within the desired tolerance $\epsilon$ and returns the current estimate of the global maximum. Otherwise, the process continues by splitting the Bernstein polynomial, using the de Casteljau algorithm (see Property \ref{prop:decast}), at an intermediate value $t_{div} = \frac{i_{ub}}{n}(t_f - t_0) + t_0$. The resulting sets of Bernstein coefficients are $\mathcal{P}^A$ and $\mathcal{P}^B$. The algorithm proceeds by making recursive calls for each pair of split curves, until the maximum value is found and returned.

\begin{remark}
By returning $\alpha$ in the third `\textit{if}' statement, we acknowledge that the true maximum value can be as much as $\epsilon$ smaller than the estimate. If a conservative estimate is preferable, one should return $\mathcal P_{ub}$ instead.
\end{remark}

\begin{algorithm}\small
\caption{Evaluating Maximum for 1-Dimensional Bernstein Polynomial}
\label{algorithm:extrema}
\normalem
\SetKwInOut{Input}{Input}
\Input{$\mathcal{P}, \alpha, \epsilon$}

    $P_{lb} = $ lower\_bound($\mathcal P$)

    $P_{ub} = \max (\mathcal P)$

    $i_{ub} = \argmax (\mathcal P)$

    \If{$P_{lb} > \alpha$}{
        $\alpha = P_{lb}$
    }

    \If{$\alpha > \mathcal P_{ub}$}{
        \Return $\alpha$
    }

    \uIf{$P_{ub} - P_{lb} < \epsilon$}{
        \Return $\alpha$
    }
    \Else{

        $\mathcal{P}^{A}, \mathcal{P}^{B} =
        \text{split}(\mathcal{P}, i_{ub})$

        $A_{\max} =$ Algorithm 1($\mathcal{P}^{A}, P_{\max}, \epsilon$)

        $B_{\max} =$ Algorithm 1($\mathcal{P}^{B}, P_{\max}, \epsilon$)

        $P_{\max} = \max(\mathcal{A}_{\max}, \mathcal{B}_{\max})$

    }

    \Return $P_{\max}$
\end{algorithm}

\begin{algorithm}\small
\caption{Minimum Distance Between Two Bernstein Polynomials}
\label{algorithm:spatSep}
\normalem
\SetKwInOut{Input}{Input}
\Input{$\mathcal{P}, \mathcal{Q}, \alpha, \epsilon$}

    $upper =$ upper\_bound($\mathcal{P}, \mathcal{Q}$)

    $lower =$ lower\_bound($\mathcal{P}, \mathcal{Q}$)

    \If{$upper < \alpha$}{

        $\alpha = upper$
    }

    \eIf{$upper - lower < \epsilon$}{
        \Return $\alpha$
    }{

        $\mathcal{P}^A, \mathcal{P}^B =$ split($\mathcal{P}$)

        $\mathcal{Q}^A, \mathcal{Q}^B =$ split($\mathcal{Q}$)

        $\alpha =$ min($\alpha$,
            Algorithm \ref{algorithm:spatSep}
            ($\mathcal{P}^A, \mathcal{Q}^A, \alpha$))

        $\alpha =$ min($\alpha$,
            Algorithm \ref{algorithm:spatSep}
            ($\mathcal{P}^A, \mathcal{Q}^B, \alpha$))

        $\alpha =$ min($\alpha$,
            Algorithm \ref{algorithm:spatSep}
            ($\mathcal{P}^B, \mathcal{Q}^A, \alpha$))

        $\alpha =$ min($\alpha$,
            Algorithm \ref{algorithm:spatSep}
            ($\mathcal{P}^B, \mathcal{Q}^B, \alpha$))
    }
\Return $\alpha$
\end{algorithm}

Algorithm \ref{algorithm:extrema} (and its converse) is employed to find the minimum and maximum of the 5th degree Bernstein polynomial depicted in Figure \ref{fig:elevminmaxhull} (red lines). The execution time to compute the minimum is 320 $\mu$s on a Lenovo ThinkPad laptop using an Intel Core i7-8550U, 1.80 GHz CPU. The implementation can be found in \cite{BeBOT}.

\subsection{Minimum Spatial Distance}
\label{sec:mindist}
The minimum spatial distance between two Bernstein polynomials can be computed using the method outlined in \cite{chang2011computation}. This is done by exploiting the Convex Hull (Property \ref{prop:convexhull}) and End Point Values (Property \ref{prop:endpts}) properties and the de Casteljau (Property \ref{prop:decast}) and Gilbert-Johnson-Keerthi (GJK) algorithms \cite{gilbert1988fast}. The latter is widely used in computer graphics to compute the minimum distance between convex shapes. The algorithm for minimum distance computation between two Bernstein polynomials is presented in Algorithm \ref{algorithm:spatSep}.

\begin{figure}
    \centering
    \includegraphics[scale=0.19]{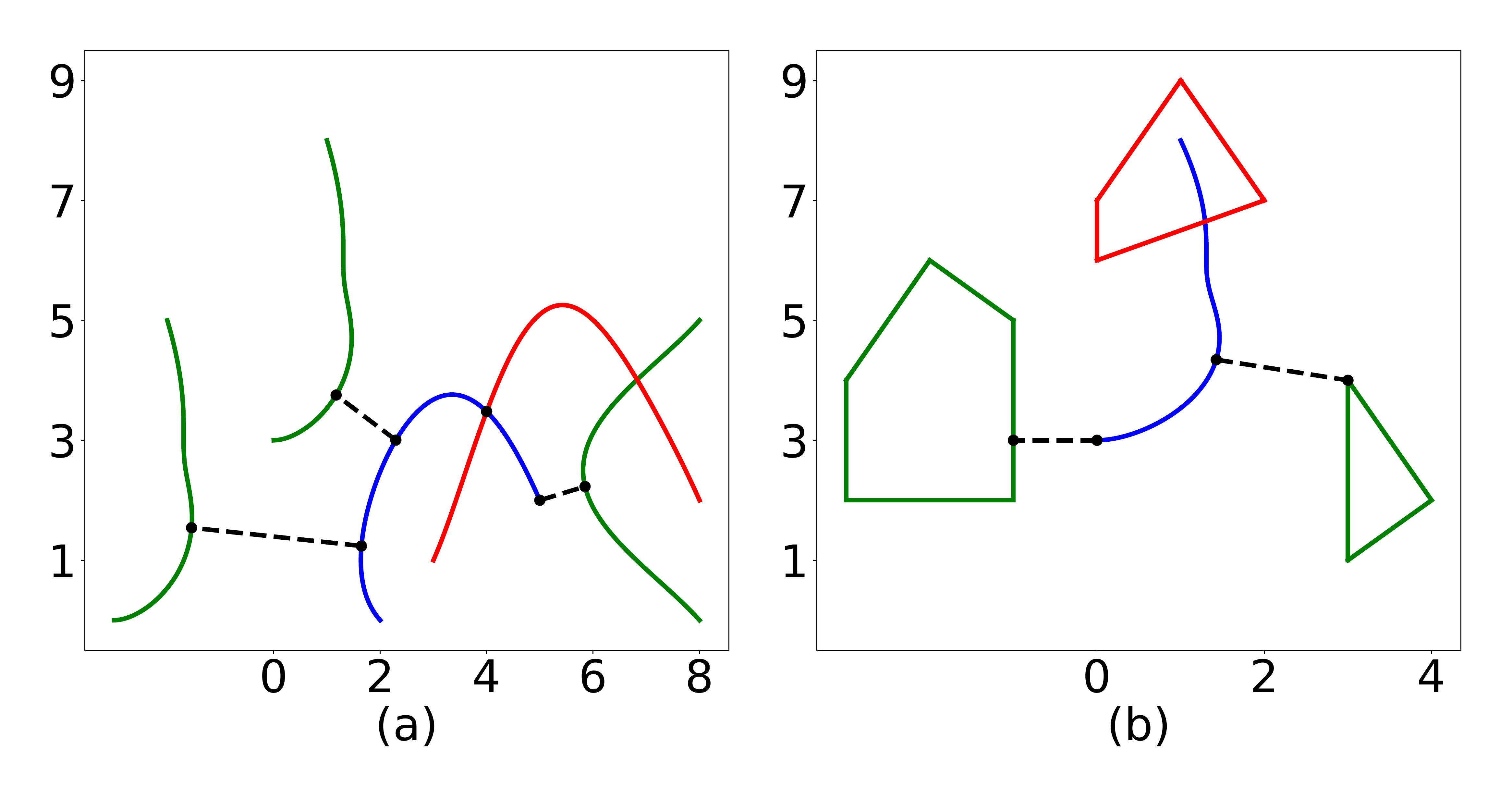}
    \caption{(a) Minimum distance between curves. (b) Minimum distance between a curve and a polygon. All distances are measured to the blue curve. A red curve or polygon indicates that a collision exists.
    }
    \label{fig:bez2bez2poly}
\end{figure}

The inputs to the minimum distance algorithm, Algorithm~\ref{algorithm:spatSep}, are the sets containing the coefficients of the Bernstein polynomials, $\mathcal{P}=\{\mathbf{P}_n\}$ and $\mathcal{Q}=\{\mathbf{Q}_n\}$, a global maximum $\alpha$, and a tolerance $\epsilon$. Similar to the $lower\_bound$ function in Sec. \ref{sec:evaluatingextrema}, the $upper\_bound$ function finds the maximum distance between the end points of the two polynomials passed in, which is the upper bound since the minimum distance between the two Bernstein polynomials will not be larger than that value (refer to Property \ref{prop:endpts}). The $lower\_bound$ function (not to be confused with the $lower\_bound$ function in Algorithm \ref{algorithm:extrema}) uses the GJK algorithm to find the distance between the convex hull of the Bernstein coefficients of the two Bernstein polynomials, which is a lower bound due to Property \ref{prop:convexhull}. Finally, the curves are split, and recursive calls are made for each pair of split curves, until the difference between the upper and lower bounds satisfies a given tolerance. Figure \ref{fig:bez2bez2poly} (a) illustrates the minimum distance between several different Bernstein polynomials. The code to generate this plot can be found in \cite{BeBOT}.

\begin{remark} \label{rem:minsep}
Note that Algorithm \ref{algorithm:spatSep} can also be employed to compute the minimum distance between a Bernstein polynomial and a point or a convex shape. This is shown in Figure \ref{fig:bez2bez2poly} (b).
\end{remark}


\subsection{Collision Detection}
The collision detection algorithm is similar, in principle, to the minimum distance algorithm. The difference between the two is instead of having the GJK algorithm return a minimum distance, it simply returns whether a collision has been detected, i.e., two convex hulls intersect. Thus, this algorithm is computationally inexpensive with respect to the minimum distance algorithm, but it returns a binary value (no collision/collision possible) rather than the actual distance between the polynomials.

The collision detection algorithm is outlined in Algorithm~\ref{algorithm:collisionDet}. The inputs to Algorithm~\ref{algorithm:collisionDet} are the sets containing the Bernstein coefficients of the Bernstein polynomials, $\mathcal{P}=\{\mathbf{P}_n\}$ and $\mathcal{Q}=\{\mathbf{Q}_n\}$, and a maximum iteration value, $max\_iter$. The $find\_collisions$ function returns two sets, $\mathcal{P}_{col}$ and $\mathcal{Q}_{col}$, which represent convex hulls of one set that were detected to collide with at least one of the convex hulls in the other set. If no collisions are found, the algorithm returns no collision found. For all the curves where a collision was found, they are split at $t = \frac{t_f-t_0}{2}+t_0$. The previous curve is removed from the set and the new curves from the split are added to the set. If the maximum number of iterations has been met, the algorithm returns that a collision is possible. The implementation of the collision detection algorithm can be found in \cite{BeBOT}.

\begin{algorithm}
\caption{Collision Detection}
\label{algorithm:collisionDet}
\normalem
\SetKwInOut{Input}{Input}
\Input{$\mathcal{P}, \mathcal{Q}, max\_iter$}

    $k = 0$

    \While{$k < max\_iter$}{

        $\mathcal{P}_{col}, \mathcal{Q}_{col} =$ find\_collisions($\mathcal{P}, \mathcal{Q}$)

        \If {$\mathcal{P}_{col} \cup \mathcal{Q}_{col} = \{\}$}{
            \Return No Collision
        }

        \For {$\mathcal{P}_i \in \mathcal{P}_{col}$}{

            $\mathcal{P}^A, \mathcal{P}^B =$ split($\mathcal{P}_i$)

            $\mathcal{P} = \mathcal{P} \cup \{\mathcal{P}^A, \mathcal{P}^B\} \setminus \mathcal{P}_i$
        }

        \For {$\mathcal{Q}_i \in \mathcal{Q}_{col}$}{

            $\mathcal{Q}^{A}, \mathcal{Q}^{B} =$ split($\mathcal{Q}_i$)

            $\mathcal{Q} = \mathcal{Q} \cup \{\mathcal{Q}^A, \mathcal{Q}^{B}\} \setminus \mathcal{Q}_i$
        }

        $k++$
    }
\Return Collision Possible

\end{algorithm}

\section{Numerical Examples}
\label{sec:examples}
In this section, numerical examples using the toolkit and Python's Scipy Optimization package are examined (flight tests are available at \cite{youtubeDemo}). The implementation of the following examples can be found in \cite{BeBOT}.

\subsection{Dubin's Car - Time Optimal} \label{sec:dubins}
In this simple example several trajectories for a vehicle with Dubin's car dynamics are generated to illustrate the properties of Bernstein polynomials. Let the position of the vehicle under consideration be parameterized as a Bernstein polynomial, namely,
\begin{equation} \label{eq:posproblem1}
    \begin{bmatrix}
        {C_{n}^{[x]}}(t) \\
        {C_{n}^{[y]}}(t)
    \end{bmatrix} = \mathbf{C}_n(t) = \sum_{i=0}^n \mathbf{P}_{i,n} B_{i,n}(t) , \quad t \in [t_0, t_f].
\end{equation}
The square of the speed of the vehicle is a 1-D Bernstein polynomial given by $$v^2(t)=||\dot{\mathbf{C}}_{n}(t)||^2.$$ The heading angle is $$\psi(t)=\tan^{-1}\frac{\dot C^{[y]}_n(t)}{\dot C^{[x]}_n(t)},$$ and the angular rate is a 1-D rational Bernstein polynomial given by
$$\omega(t)=\frac{\ddot C^{[y]}_n(t)\dot C^{[x]}_n(t)-\dot C^{[y]}_n(t)\ddot C^{[x]}_n(t)}{||\dot{\mathbf{C}}_{n}(t)||^2}.$$
The goal is to compute a time optimal trajectory subject to maximum speed and angular rate bounds, initial and final position, angle, and speed. The vehicle must also maintain a minimum safe distance from two obstacles. This problem can be formulated as follows:
\begin{equation*}
	\begin{split}
		& \min_{\mathbf{P}_n, t_f} t_f
	\end{split}
\end{equation*}
subject to
\begin{equation*}
	\begin{split}
		& \mathbf{C}_n(0) = \mathbf{C}_0, \quad
		  \mathbf{C}_n(t_f) = \mathbf{C}_f, \\
		& \psi(0) = \psi_0, \quad \psi(t_f) = \psi_f \\
		& ||\dot{\mathbf{C}}_n(0)|| = v_0, \quad
		  ||\dot{\mathbf{C}}_n(t_f)|| = v_{f}, \\
		& ||\dot{\mathbf{C}}_n(t)||^2 \leq v^2_{\max}, \quad
		  \forall t \in [0,t_f]\\
		& ||\dot{\psi}(t)|| \leq \omega_{\max}, \quad
		  \forall t \in [0,t_f] \\
		& ||\mathbf{C}_n(t) - \mathbf{O}_i||^2 \geq d_s^2, \quad
		  \forall t \in [0,t_f], \quad i = 1,2.
	\end{split}
\end{equation*}
where $\mathbf{O}_i$ is the position of the $i$th obstacle.

The initial and final positions are $\mathbf{C}_0 = [3,0]^\top m$ and $\mathbf{C}_f = [7,10]^\top m$, respectively. The initial and final headings are $\psi_0 = \psi_f = \frac{\pi}{2} rad$ (facing toward the positive $Y$ direction). The initial and final speeds are given by $v_0=v_f=1 \frac{m}{s}$. The maximum speed and angular rate constraints are $v_{\max}=5 \frac{m}{s}$ and $\omega_{\max}=1 \frac{rad}{s}$, respectively.


In the problem above, the order of the Bernstein polynomial, $\mathbf{C}_n (t)$, is set to $n=10$. The initial and final position constraints are enforced using the End Point Values property (Property \ref{prop:endpts}). Similarly, the same property is used to enforce the initial and final speeds and headings (see \eqref{eq:derinitpoint}). Note that the norm squared of the speed and of the distance between the trajectory and the obstacles can be expressed as $1$-dimensional Bernstein polynomials (the sum, the difference, and the product between Bernstein polynomials are also Bernstein polynomials). A similar argument can be made for the norm square of the angular rate, which can be expressed as a rational Bernstein polynomial (see Property \ref{prop:arithmetic}). Thus, the maximum speed and angular rate, and collision avoidance constraints can be enforced using the Evaluating Bounds or Evaluating Extrema procedures described in Sections \ref{sec:evaluatingbounds} and \ref{sec:evaluatingextrema}.

\begin{figure}
	\centering
	\includegraphics[scale=0.25]{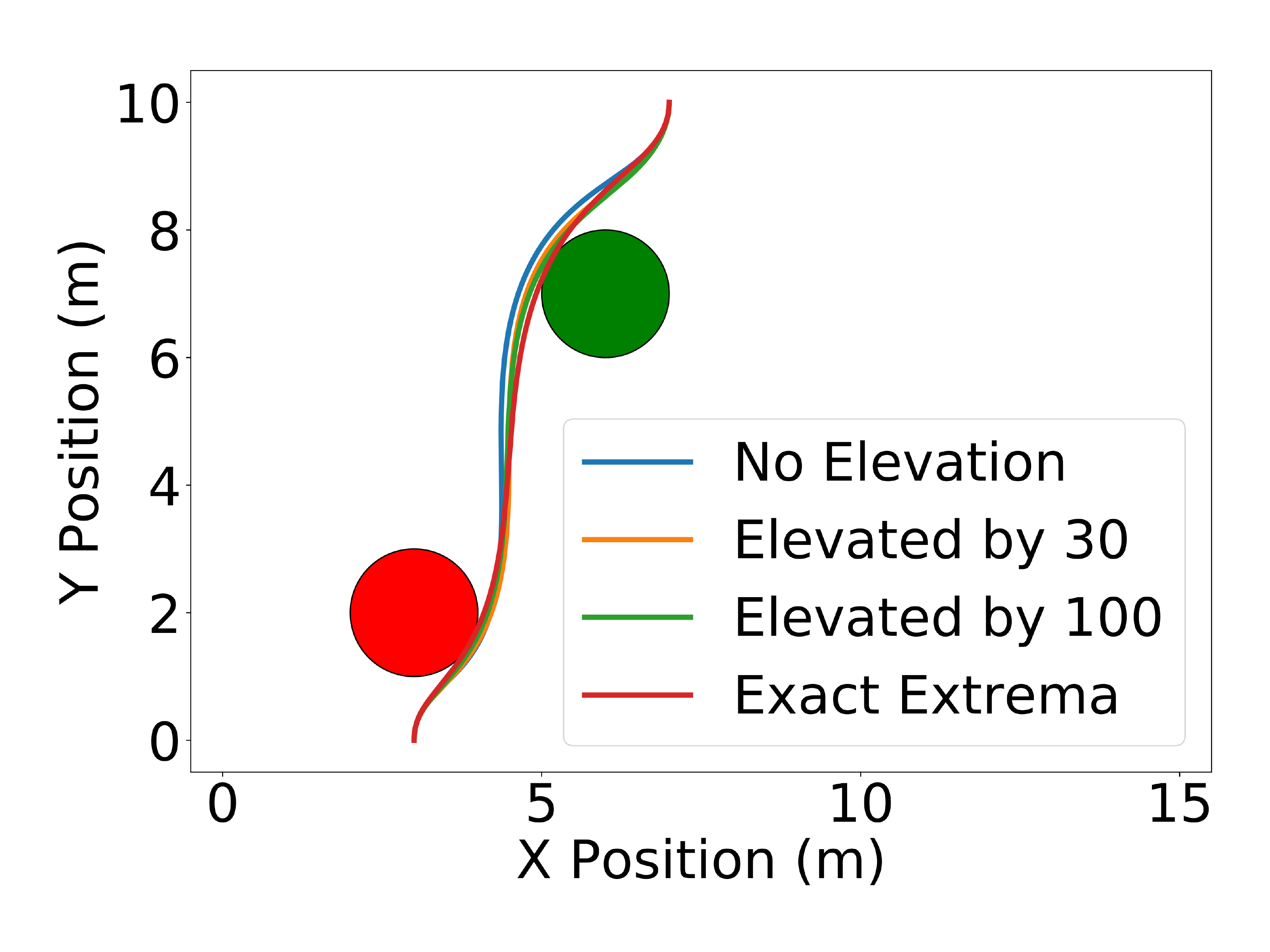}
	\caption{Time optimal trajectory for vehicle with initial and final speeds and headings, maximum speed, maximum angular rate, and maximum safe distance constraints ranging from least to most conservative distance estimates.}
	\label{fig:timeOpt}
\end{figure}

In Figure \ref{fig:timeOpt}, the blue curve is obtained by enforcing the constraints using the Evaluating Bounds procedure without degree elevation. The optimal time for the procedure without degree elevation is $t_f = 9.14 s$. The orange and green curves are obtained by enforcing the constraints using the same property after degree elevations of $30$ and $100$, respectively. Degree elevation to degree $30$ results in an optimal final time $t_f = 7.64 s$. The elevation to degree $100$ provides an optimal value $t_f = 7.12 s$. Finally, the trajectory with smallest optimal final time, $t_f = 6.45 s$, depicted as the red curve in Figure \ref{fig:timeOpt}, is obtained by enforcing the constraints using the Evaluating Extrema algorithm (Section \ref{sec:evaluatingextrema}).

\begin{figure}
	\centering
	\includegraphics[scale=0.27]{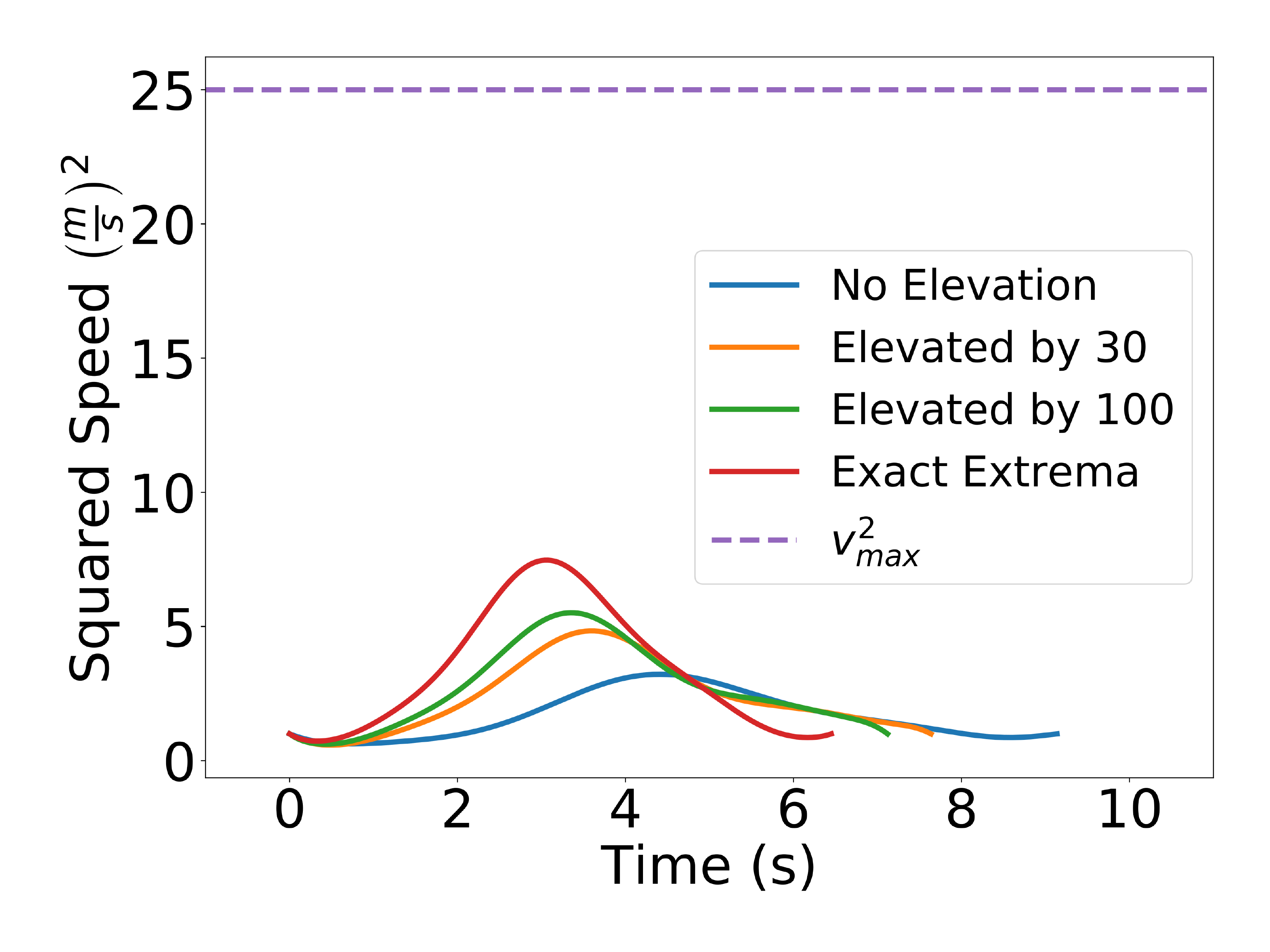}
	\caption{Plot of the squared speed constraints for each separate trial.}
	\label{fig:dubSpeed}
\end{figure}

\begin{figure}
	\centering
	\includegraphics[scale=0.27]{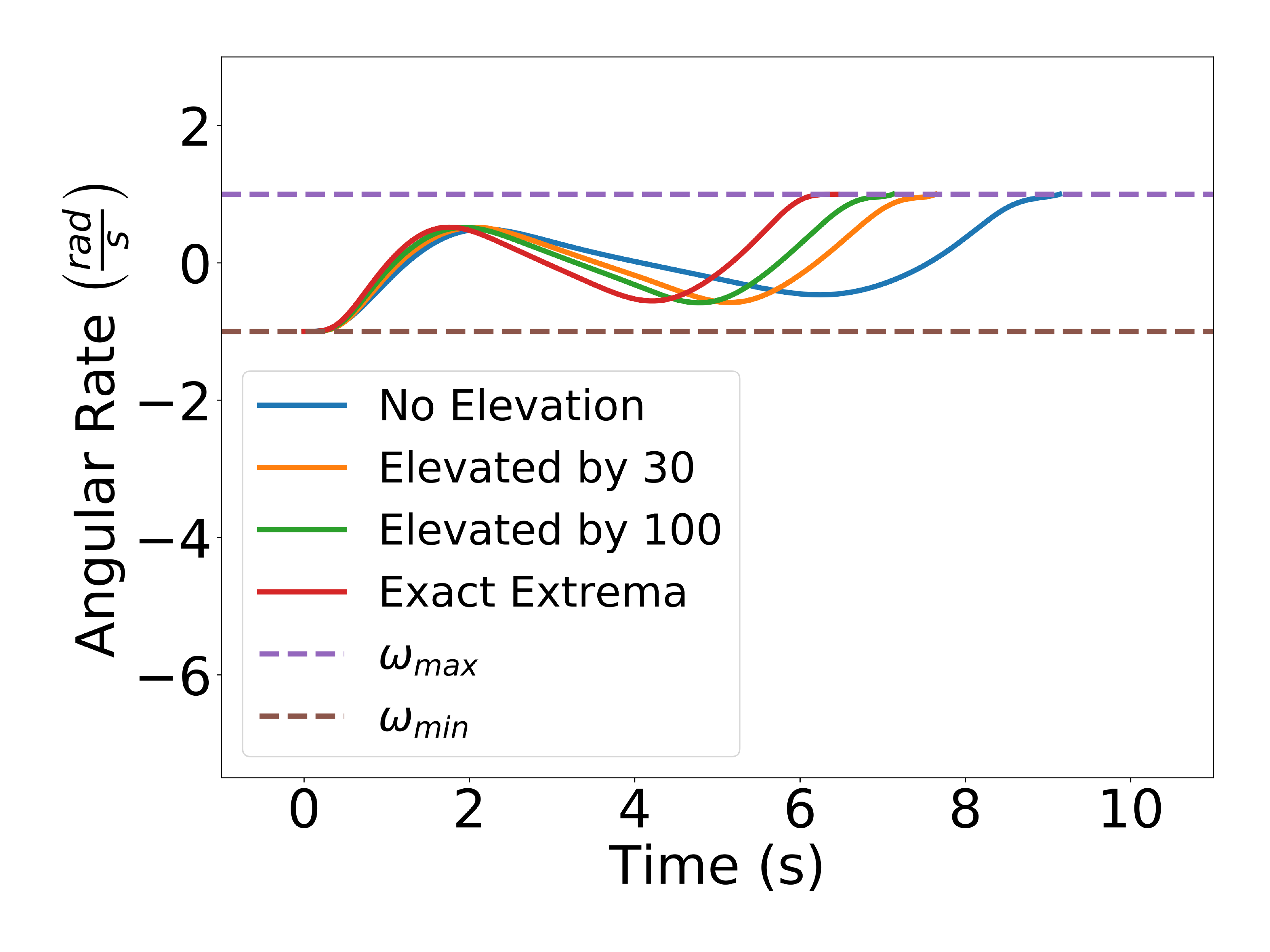}
	\caption{Plot of the angular rate constraints for each separate trial.}
	\label{fig:dubAngRate}
\end{figure}

Figure \ref{fig:dubSpeed} illustrates the squared speed of each example. As the trajectory is able to pass closer to the obstacle, the maximum speed is increased and in turn lowers the final time.
Figure \ref{fig:dubAngRate} shows the angular rate of each trial. It can be seen that the vehicle correctly adheres to the angular rate constraints for each trial with the only differences being the final time and proximity to the obstacles.

\begin{remark}
The Exact Extrema function is a complex non-linear and non-smooth function. When it is used to enforce constraints, gradient-based optimization solvers such as the one used in this work can fail to converge to a feasible solution, especially if the initial guess is not feasible. One option is to use an iterative procedure where (1) a feasible sub-optimal solution is obtained by enforcing the collision avoidance constraint using the Evaluating Bounds function, and (2) this solution is then used as an initial guess to solve the (more accurate) problem with the Exact Extrema constraint.
\end{remark}

\subsection{Air Traffic Control - Time Optimal} \label{sec:atc}
In this example we consider the problem of routing several commercial flights between major US cities in two dimensions (i.e. constant altitude). Assuming that each flight departs at the same time, the goal is to minimize the combined flight time of all the vehicles. Let the position, speed, heading, and angular rate of each vehicle under consideration be parameterized as in Section \ref{sec:dubins}. We shall also make the assumption that the trajectories are on a 2D plane rather than on the surface of a sphere.

The goal is to compute cumulatively time optimal trajectories subject to maximum speed and angular velocity bounds, initial and final position, angle, and speeds. The vehicles must also maintain a minimum safe distance between each other. This problem can be formulated as follows:
\begin{equation*}
	\min_{\mathbf P_n, \mathbf t_f} \sum_{k=1}^{m} t^{[k]}_f
\end{equation*}
subject to
\begin{equation*}
	\begin{split}
		& \mathbf{C}^{[k]}_n(0) = \mathbf{C}^{[k]}_0, \quad
		  \mathbf{C}^{[k]}_n\left(t_f^{[k]}\right) = \mathbf{C}^{[k]}_f, \\
		& \psi^{[k]}(0) = \psi^{[k]}_0, \quad \psi^{[k]}\left(t_f^{[k]}\right) = \psi^{[k]}_f, \\
		& ||\dot{\mathbf{C}}^{[k]}_n(0)|| = v^{[k]}_0, \quad
		  ||\dot{\mathbf{C}}^{[k]}_n\left(t_f^{[k]}\right)|| = v^{[k]}_{f}, \\
		& v^2_{\min} \leq ||\dot{\mathbf{C}}^{[k]}_n(t)||^2 \leq v^2_{\max}, \quad
		  \forall t \in [0,t_f^{[k]}], \\
		& ||\dot{\psi}(t)|| \leq \omega_{\max}, \quad
		  \forall t \in [0,t^{[k]}_f], \\
		& ||\mathbf{C}_n^i(t) - \mathbf{C}_n^j(t)||^2 \geq d_s^2, \quad \forall i,j \in \{1, \dots, m\}, i \neq j.
	\end{split}
\end{equation*}
where the superscript $[k]$ corresponds to the $k$th vehicle out of $m$ vehicles, $\mathbf C^{[k]}_0$ and $\mathbf C^{[k]}_f$ are the initial and final positions, $\psi^{[k]}_0$ and $\psi^{[k]}_f$ are the initial and final headings, $v^{[k]}_0$ and $v^{[k]}_f$ are the initial and final speeds, $v_{\min}$ and $v_{\max}$ are the minimum and maximum speeds, $\omega_{\max}$ is the maximum angular velocity, $d_s$ is the minimum safe distance, and $t_f^{[k]}$ is the final time of the $k$th vehicle.

The departure cities, in vehicle order, are: San Diego, New York, Minneapolis, and Seattle. The arrival cities, in vehicle order, are: Minneapolis, Seattle, Miami, and Denver. The initial and final speeds are all $v^{[k]}_0 = v^{[k]}_f = 205 \frac{m}{s} \, \forall k \in \{1,\dots,m\}$, the initial headings are $\mathbf \psi_0 = [0, \pi, 0, 0]^\intercal rad$, the final headings are $\mathbf \psi_f = [0, \pi, -\frac{\pi}{2}, 0] rad$, the minimum speed is $v_{\min} = 200 \frac{m}{s}$, the maximum speed is $v_{\max} = 260 \frac{m}{s}$, the maximum angular velocity is $\omega_{\max} = 3 \frac{deg}{s} = 0.0524 \frac{rad}{s}$, the minimum safe distance is $d_s = 5 km$, and the degree of the Bernstein polynomials being used is $5$.

The initial and final position constraints are enforced using the End Point Values property (Property \ref{prop:endpts}). Similarly, the same property is used to enforce the initial and final speeds and headings (see \eqref{eq:derinitpoint}). Note that the norm square of the speed and the norm square of the distance between vehicles can be expressed as $1$-dimensional Bernstein polynomials (the sum, difference, and product between Bernstein polynomials are also Bernstein polynomials). A similar argument can be made for the norm square of the angular rate, which can be expressed as a rational Bernstein polynomial (see Property \ref{prop:arithmetic}). Thus, the maximum speed and angular rate, and collision avoidance constraints can be enforced using the Evaluating Bounds or Evaluating Extrema procedures described in Sections \ref{sec:evaluatingbounds} and \ref{sec:evaluatingextrema}.

\begin{figure}
	\centering
	\includegraphics[scale=0.27]{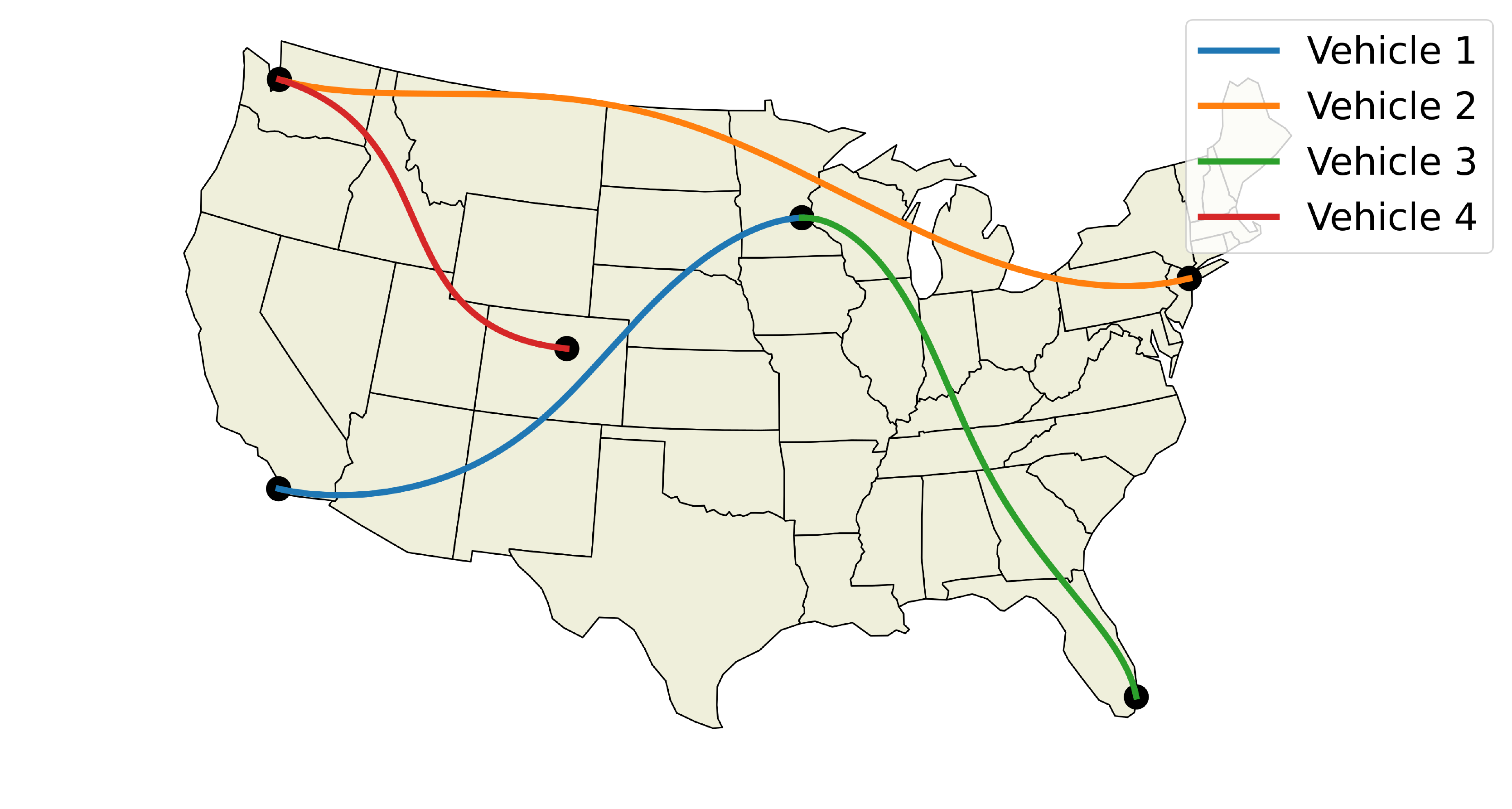}
	\caption{Commercial flight trajectories between major US cities.}
	\label{fig:atc}
\end{figure}

\begin{figure}
	\centering
	\includegraphics[scale=0.27]{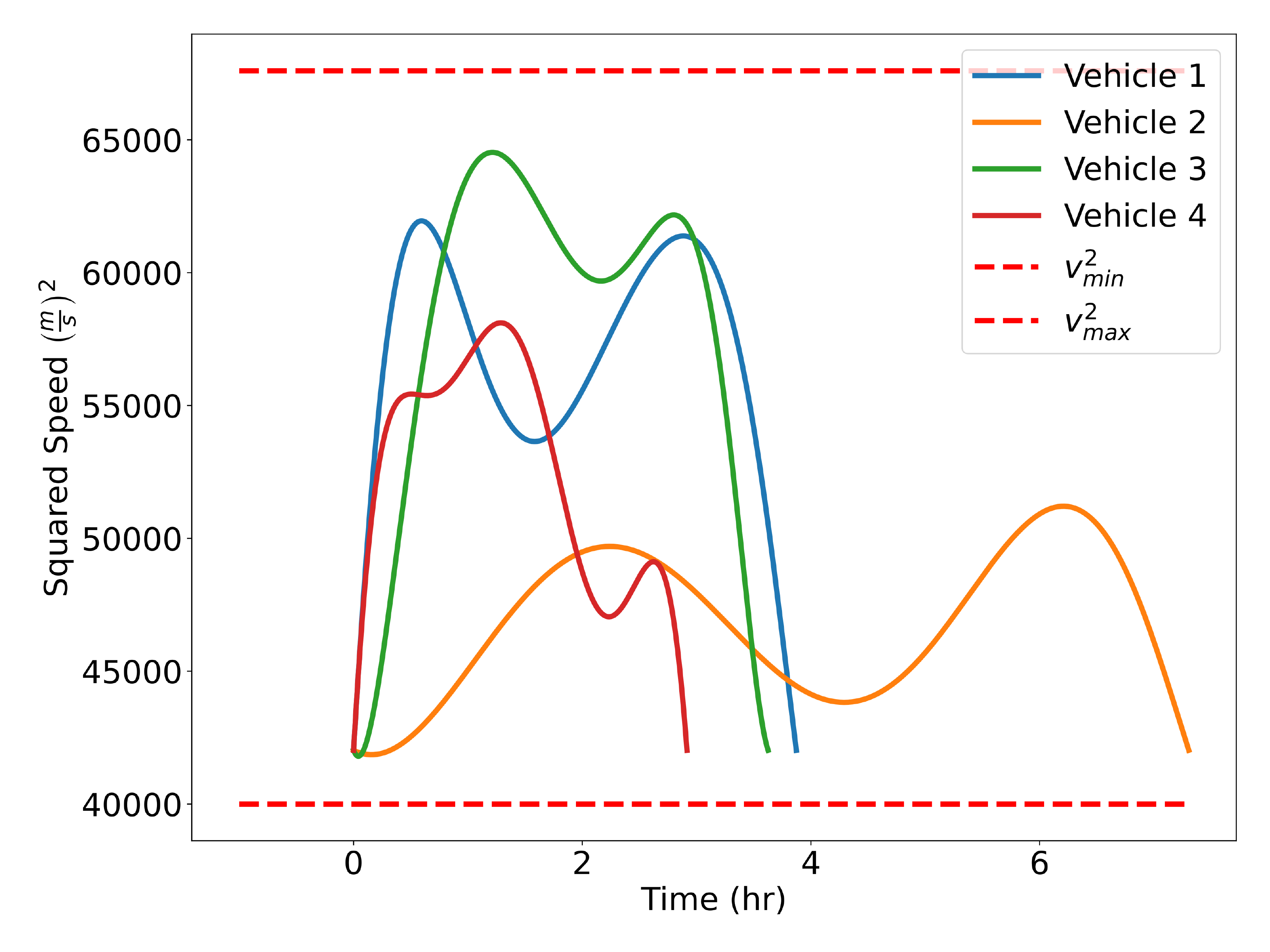}
	\caption{Verifying speed constraints for the Air Traffic Control example.}
	\label{fig:atc_speeds}
\end{figure}

\begin{figure}
	\centering
	\includegraphics[scale=0.27]{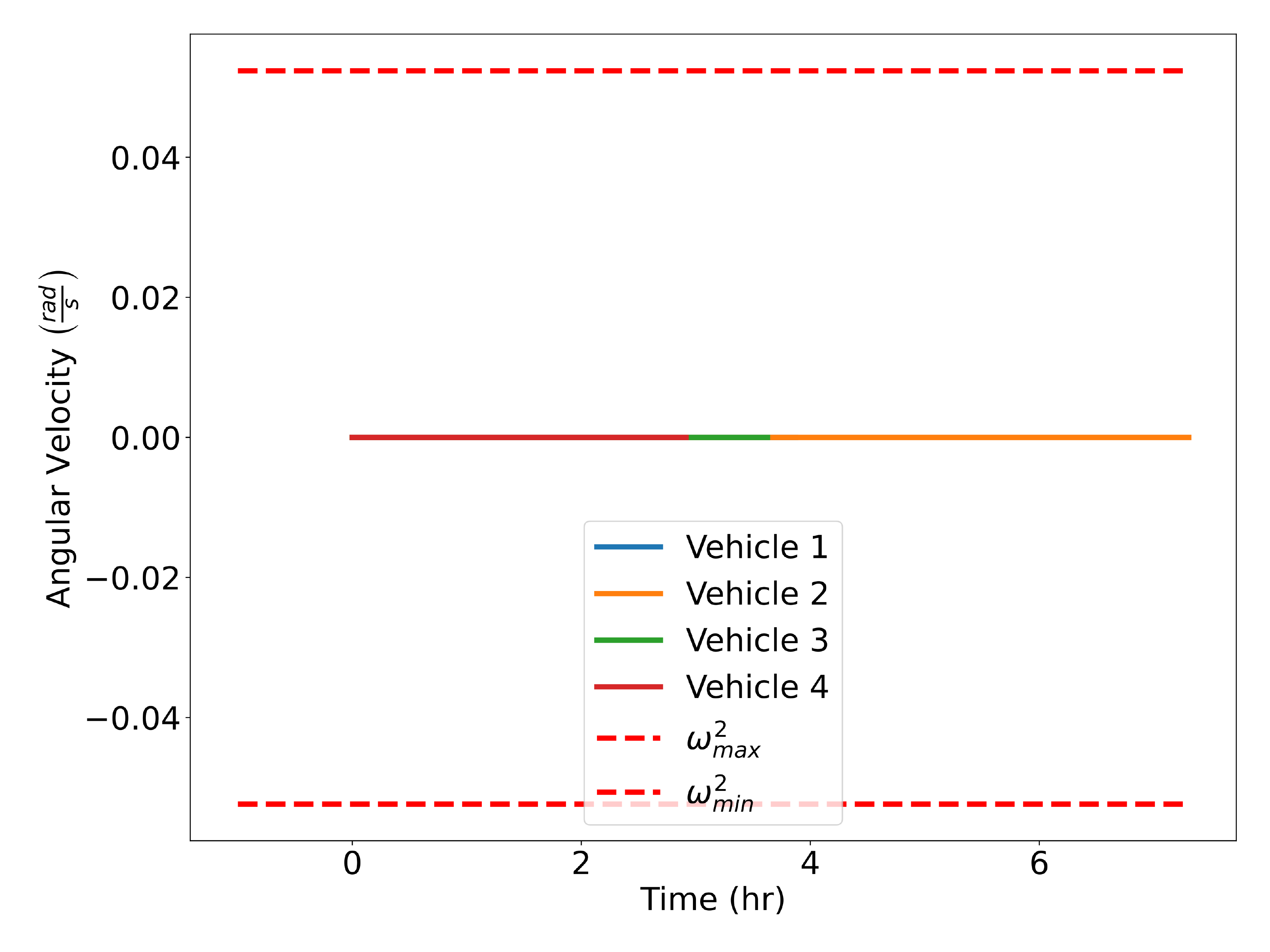}
	\caption{Verifying angular rate constraints for the Air Traffic Control example.}
	\label{fig:atc_angrates}
\end{figure}

\begin{figure}
	\centering
	\includegraphics[scale=0.27]{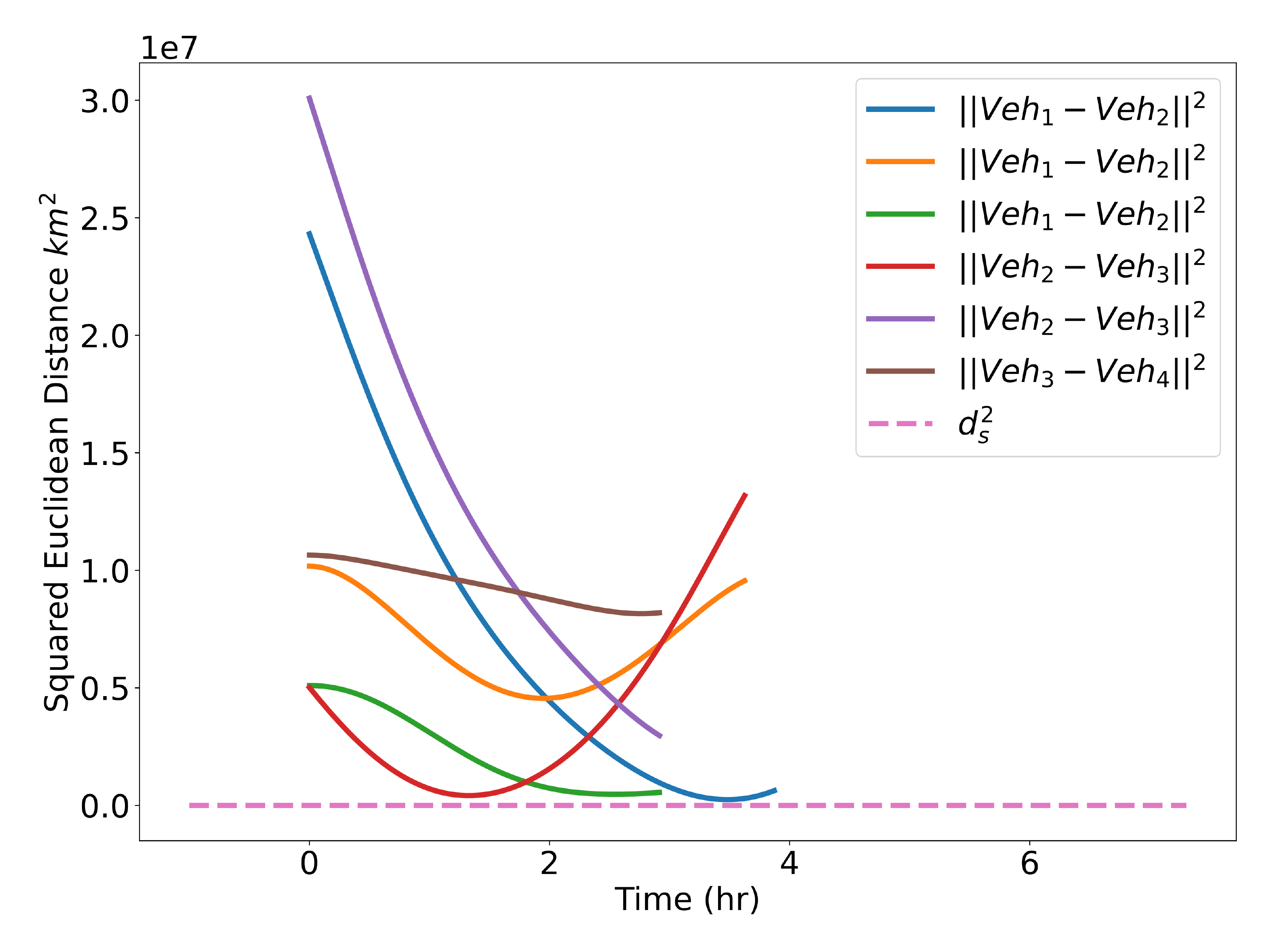}
	\caption{Verifying minimum safe distance constraints for the Air Traffic Control example.}
	\label{fig:atc_distances}
\end{figure}

The optimized flight plans can be seen in Figure \ref{fig:atc}. The squared speed of each vehicle is shown in Figure \ref{fig:atc_speeds}. Note that each vehicle begins and ends with the same speed. This is due to the initial and final speed constraints of the problem. The vehicles never slow down less than their initial speeds which means they never reach the minimum speed constraint. Nor do the vehicles go faster than the maximum speed. In Figure \ref{fig:atc_angrates}, the angular velocity of each vehicle is shown. The minimum and maximum angular rate constraints are shown by the dotted lines. The vehicles' angular rates never approach the minimum or maximum angular rate constraints due to the large area being covered. Finally, the squared euclidean distance between vehicles is shown in Figure \ref{fig:atc_distances}. As expected, the squared Euclidean distance between two vehicles never falls below the minimum safe distance. Note that curves within the constraint plots end at different times. This is expected since each vehicle has a different final time. The furthest time reached in Figure \ref{fig:atc_distances} is less than that of the other plots because the other vehicles have already reached their final time before the longest flight reaches its final time.

\subsection{Cluttered Environment}
In many real world scenarios robots must safely traverse cluttered environments. In this example, three aerial vehicles traveling at a constant altitude must navigate around several obstacles while also adhering to dynamic and minimum safe distance constraints. Let the position, speed, heading angle, and angular rate of each vehicle be defined as in Section \ref{sec:dubins}. The goal of this example is to compute trajectories whose arc length is minimized subject to maximum speed constraints along with initial and final positions, heading angles, and speeds. The vehicles should also adhere to a minimum safe distance between each other and between obstacles. We formulate the problem as follows:
\newpage
\begin{equation}
	\min_{\mathbf P_n} \sum_{i=1}^m \sum_{k=0}^{n-1} ||\mathbf P_{k+1}^{[i]} - \mathbf P_k^{[i]}||
\end{equation}
subject to
\begin{equation*}
	\begin{split}
		& \mathbf{C}^{[k]}_n(0) = \mathbf{C}^{[k]}_0, \quad
		  \mathbf{C}^{[k]}_n(t_f) = \mathbf{C}^{[k]}_f, \\
		& \psi^{[k]}(0) = \psi^{[k]}_0, \quad \psi^{[k]}(t_f) = \psi^{[k]}_f, \\
		& ||\dot{\mathbf{C}}^{[k]}_n(0)|| = v^{[k]}_0, \quad
		  ||\dot{\mathbf{C}}^{[k]}_n(t_f)|| = v^{[k]}_{f}, \\
		& ||\dot{\mathbf{C}}^{[k]}_n(t)||^2 \leq v^2_{\max}, \quad
		  \forall t \in [0,t_f], \\
		& ||\mathbf{C}_n^{[i]}(t) - \mathbf{C}_n^{[j]}(t)||^2 \geq d_s^2, \quad \forall i,j \in \{1, \dots, m\}, i \neq j, \\
        & ||\mathbf{C}_n^{[i]}(t) - \mathbf{O}_j||^2 \geq d_{obs}^2, \quad
          \forall t \in [0,t_f], \quad i\in \{1, \dots, m\}, \\
        & \quad \quad j \in \{1, \dots, b\}.
    \end{split}
\end{equation*}
Where $\mathbf O_j$ is the position of the $j$th obstacle out of $b$ obstacles.

\begin{figure}
	\centering
	\includegraphics[scale=0.15]{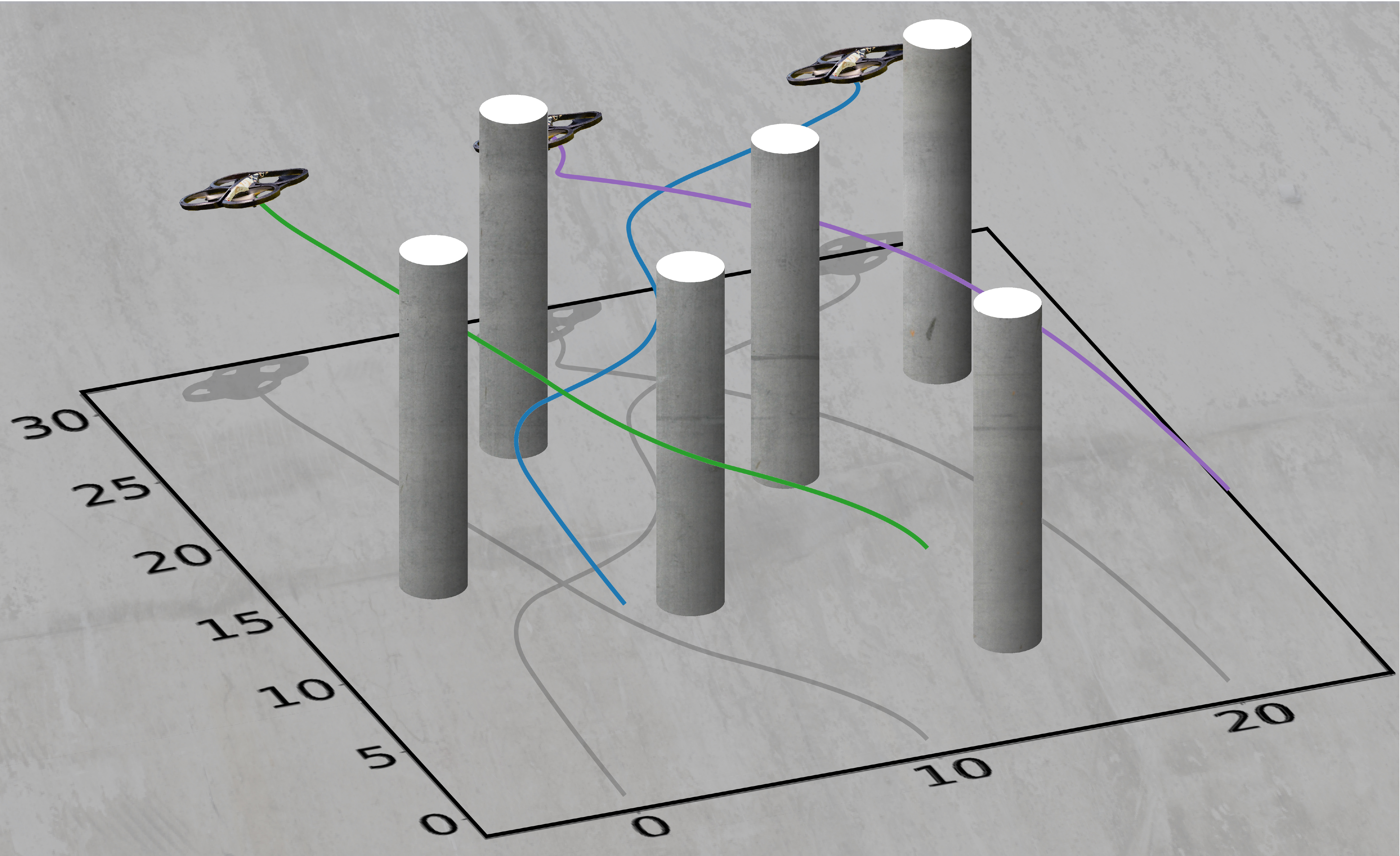}
	\caption{Aerial vehicles navigating a cluttered environment.}
	\label{fig:clutenv}
\end{figure}

The initial positions for each vehicle, in order, are $[0, 0]^\intercal m$, $[10, 0]^\intercal m$, and $[20, 0]^\intercal m$. The initial speeds are all $1 \frac{m}{s}$ and the initial heading angles are all $\frac{\pi}{2} rad$. The final positions for each vehicle are, in order, $[20, 30]^\intercal m$, $[0, 30]^\intercal m$, and $[10, 30]^\intercal m$. The final speeds and final heading angles are the same as the initial speeds and heading angles. The order of the Bernstein polynomials being used is $7$, the final time is $t_f = 30 s$, the minimum safe distance between vehicles is $d_s = 1 m$, the minimum safe distance between vehicles and obstacles is $d_{obs} = 2 m$, and the maximum speed is $v_{\max} = 10 \frac{m}{s}$. The vehicles traversing the cluttered environment can be seen in Figure \ref{fig:clutenv}. This experiment has been repeated in the Cooperative Autonomous Systems (CAS) lab using three AR 2.0 Drones. The flight tests can be found at \cite{youtubeDemo}.

\subsection{Swarming}

This section examines two methods for generating trajectories for large groups of autonomous aerial vehicles. The centralized method optimizes every trajectory at once. On the other hand, the decentralized method generates trajectories one at a time and compares them to previously generated trajectories.

The position of each vehicle in a swarm of $m$ vehicles for the following examples is parameterized as a 3-dimensional Bernstein polynomial, i.e.,

$$
\sum_{i=0}^n \mathbf{P}_{i,n}^{[j]} B_{i,n}(t)
= \mathbf{C}_n^{[j]}(t), \quad \forall j \in \{1,\ldots,m\}.
$$

\subsubsection{101 Vehicle - Centralized}

The centralized method optimizes the trajectories for each vehicle simultaneously. The goal is to minimize the arc length of each trajectory. There are $m$ vehicles with $n$th order Bernstein polynomials representing their trajectories which are constrained to a minimum safe distance between each other and initial and final positions. This is formulated as follows:

\begin{equation*}
    \min_{\mathbf P_n} \sum_{i=1}^m \sum_{k=0}^{n-1} ||\mathbf{P}_{k+1}^{[i]} - \mathbf{P}_k^{[i]} ||,
\end{equation*}
subject to
\begin{equation*}
    \begin{split}
        & \mathbf{C}_n^{[i]}(0) = \mathbf{C}_0^i, \quad
          \mathbf{C}_n^{[i]}(t_f) = \mathbf{C}_f^{[i]}, \quad \forall i \in \{1, \dots, m\}, \\
        & ||\mathbf{C}_n^{[i]}(t) - \mathbf{C}_n^{[j]}(t)||^2 \geq d_s^2, \quad \forall i,j \in \{1, \dots, m\}, i \neq j.
    \end{split}
\end{equation*}

\begin{figure}
    \centering
    \includegraphics[trim=200 0 0 0, scale=0.3]{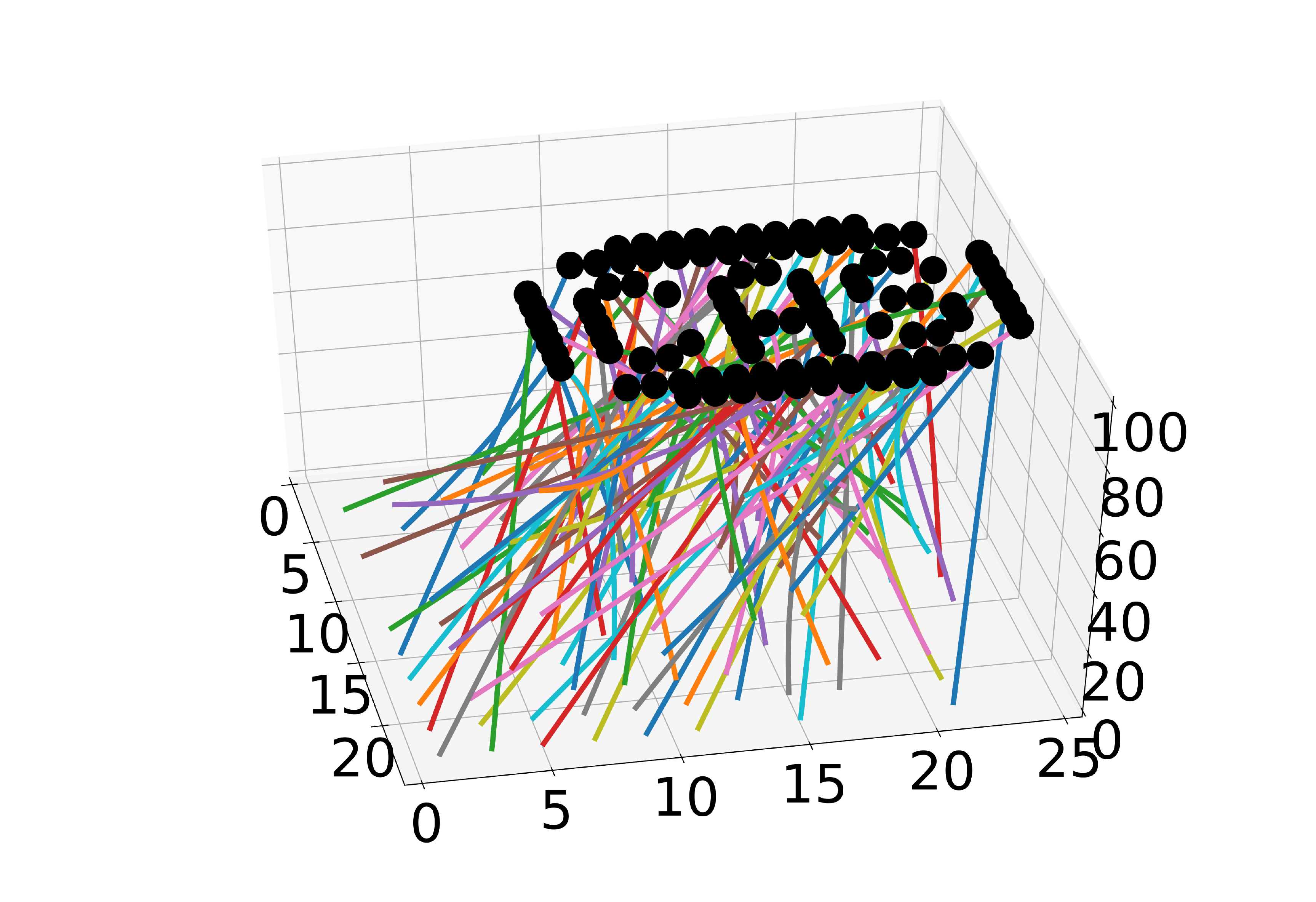}
    \caption{101 vehicles spelling out CAS using the centralized method.}
    \label{fig:101vehoptimal}
\end{figure}

The initial positions for each vehicle were chosen randomly from a $25 m \times 25 m$ grid at an altitude of $z = 0 m$. The final positions were chosen to spell out "CAS", as seen in Figure \ref{fig:101vehoptimal}, at an altitude of $z = 100 m$.
In the next section we significantly reduce the number of dimensions in the optimization vector by using the decentralized approach.


\subsubsection{101 Vehicle - Decentralized}

The decentralized method iteratively computes trajectories for the $i$th vehicle. Each new iteration is compared to the previously computed trajectories so that the minimum safety distance constraint is met. The problem that is solved at each iteration is written as

\begin{equation*}
    \min_{\mathbf{P}^{[i]}_n}  \sum_{i=1}^m \sum_{k=0}^{n-1} ||\mathbf{P}_{k+1}^{[i]} - \mathbf{P}_k^{[i]} ||
\end{equation*}
subject to
\begin{equation*}
    \begin{split}
        & \mathbf{C}_n^{[i]}(0) = \mathbf{C}_0^{[i]}, \quad
          \mathbf{C}_n^{[i]}(t_f) = \mathbf{C}_f^{[i]}, \\
        & ||\mathbf{C}_n^{[i]}(t) - \mathbf{C}_n^{[j]}(t)||^2 \geq D_s^2, \quad \forall j \in \{1, \dots, i-1\}, \quad i > 1.
    \end{split}
\end{equation*}

Note that the first vehicle does not need to satisfy the minimum safe distance constraint since no trajectories have been computed before it.

The parameters used in this example were identical to that of the previous subsection. The resulting figure has been omitted due to its similarity to Figure \ref{fig:101vehoptimal}.


\begin{remark}
    While the decentralized method can provide faster computation times, it is not guaranteed to find a solution. For example, if previous trajectories took up too much space, a future trajectory may not have a feasible solution. In the centralized method, this problem would not occur since the optimizer could move all the trajectories in such a way that they all remain feasible. This is a common issue when using decentralized approaches. However, decentralized approaches can offer faster computation times and are robust to certain types of network drops.
\end{remark}

\subsubsection{1000 Vehicle - decentralized}

\begin{figure}
    \centering
    \includegraphics[scale=0.29]{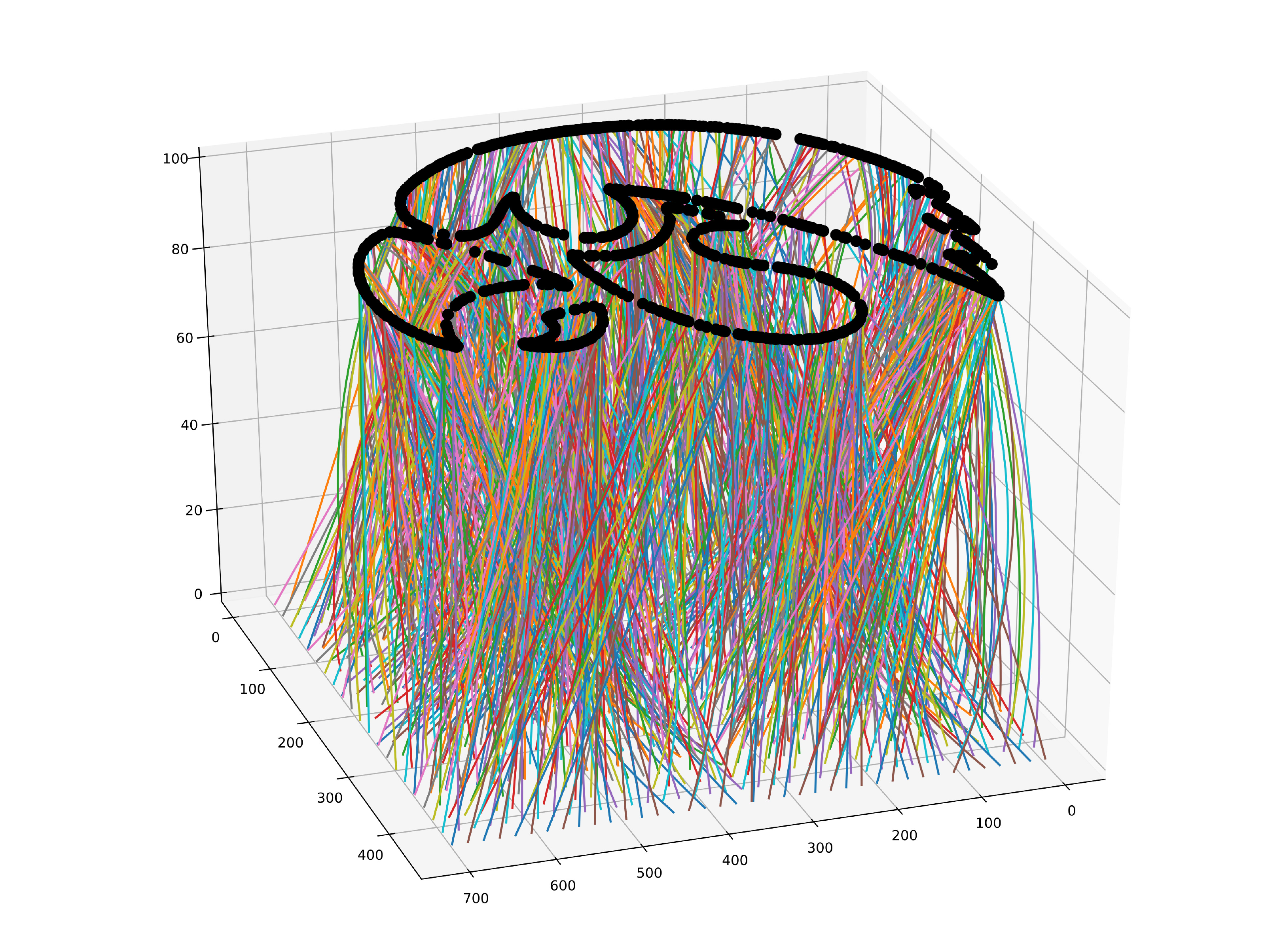}
    \caption{Trajectories for 1000 aerial vehicles with initial and final position and minimum safety distance constraints.}
    \label{fig:1000hawkslogo}
\end{figure}

The decentralized method can be used to compute 1000 trajectories. In this example, it is employed to generate the paths seen in Figure \ref{fig:1000hawkslogo} to display the University of Iowa Hawkeye logo. The initial points are equally dispersed at an altitude of $z = 0m$ on a $100m \times 100m$ grid. The final points are the pattern shown at an altitude of $z = 100$.
The cost function aims to maximize the temporal distance between the current $i$th trajectory and the previously generated $j$th trajectories by taking the reciprocal of the sum of the Bernstein coefficients of the norm squared difference, i.e.
\begin{equation*}
    \min_{\mathbf{P}^{[i]}_n} \frac{1}{\sum_{j=1}^{i-1} \mathbf{P}^{[norm, j]}}, \quad i > 1,
\end{equation*}

subject to
\begin{equation*}
    \begin{split}
        & \mathbf{C}_n^{[i]}(0) = \mathbf{C}_0^{[i]}, \quad
          \mathbf{C}_n^{[i]}(t_f) = \mathbf{C}_f^{[i]},
    \end{split}
\end{equation*}
where $\mathbf{P}^{[norm, j]}$ are the Bernstein coefficients of the Bernstein polynomial representing the squared temporal distance between the $i$th and $j$th trajectories, i.e.
$$
||\mathbf C^{[i]}(t) - \mathbf C^{[j]}(t)||^2 = \sum_{i=0}^n P^{[norm, j]} B_{i, n}(t).
$$

It should be noted that this formulation of cost function and constraints is used as a proof of concept but many other possible functions exist. For other possible cost function and constraint formulations, the reader is referred to \cite{hauser2006barrier,zhang2020optimization}.

\section{Conclusions}
\label{sec:conclusions}
We presented a method to generate optimal trajectories by using Bernstein polynomials to transcribe the problem into a nonlinear programming problem. By exploiting the useful properties of Bernstein polynomials, our method provides computationally efficient algorithms that can also guarantee safety in continuous time which are useful in optimization routines. We also developed an open source toolbox which makes these transcription methods readily available in the Python programming language.

Numerical examples were provided to demonstrate the efficacy of the method. Simple cost functions and constraints were implemented to generate trajectories which avoided obstacles, safely plan commercial flight paths, navigate a team of drones through a cluttered environment, and even generate 1000 trajectories to display a university logo. Our formulation offers a powerful tool for users to generate optimal trajectories in real time scenarios for single or multiple robot teams. Future work includes developing new cost functions, exploring different optimization frameworks, and replanning trajectories to react to a changing environment.

\bibliographystyle{IEEETran}
\bibliography{references}

\end{document}